\documentclass[letterpaper, 10pt, conference, twoside]{style/ieeeconf}    % Use this line for a4 paper

\makeatletter
\let\NAT@parse\undefined
\makeatother

\usepackage{dblfloatfix}

\usepackage[numbers,sectionbib,sort&compress]{natbib}
\usepackage{bm}
\usepackage{gensymb}
\usepackage{graphicx}
\usepackage{amsmath}
\usepackage{amssymb}
\usepackage{algorithm}
\usepackage[noend]{algpseudocode}
\usepackage{subcaption}
\usepackage{amsfonts}
\usepackage{siunitx}
\usepackage{booktabs}
\usepackage{makecell}
\usepackage{multirow}
\usepackage{upgreek}
\usepackage[font=small]{caption}
\usepackage[export]{adjustbox}
\usepackage{tikz}
\usepackage{sidecap} \sidecaptionvpos{figure}{c}
\DeclareMathOperator*{\argmax}{argmax}

\usepackage{hyperref}
\hypersetup{colorlinks,breaklinks,
linkcolor=[rgb]{0.5,0.,0.},
citecolor=[rgb]{0.000,0.427,0.173},
urlcolor=[rgb]{0.031,0.318,0.612}}

\usepackage[nameinlink, capitalize]{cleveref}
\usepackage{color}
\definecolor{CommentPink}{rgb}{1,0.2,0.5}
\definecolor{CommentBlue}{rgb}{0,0,1}
\definecolor{CommentGreen}{rgb}{0,1,0}

\Crefname{section}{Sec.}{Sec.}
\Crefname{equation}{Eq.}{Eq.}

\usepackage[printonlyused,withpage,nolist,nohyperlinks]{acronym}
\begin{acronym}

\acro{2D}{two-dimensional}

\acro{3D}{three-dimensional}

\acro{AHRS}{attitude and heading reference system}

\acro{AUV}{autonomous underwater vehicle}

\acro{CPP}{Chinese Postman Problem}

\acro{DoF}{degree of freedom}
\acrodefplural{DoF}[DoFs]{degrees of freedom}

\acro{DVL}{Doppler velocity log}

\acro{FSM}{finite state machine}

\acro{IMU}{inertial measurement unit}

\acro{LBL}{Long Baseline}

\acro{MCM}{mine countermeasures}

\acro{MDP}{Markov decision process}
\acrodefplural{MDP}[MDPs]{Markov decision processes}

\acro{POMDP}{Partially Observable Markov Decision Process}
\acrodefplural{POMDP}[POMDPs]{Partially Observable Markov Decision Processes}

\acro{PRM}{Probabilistic Roadmap}
\acrodefplural{PRM}[PRM]{Probabilistic Roadmaps}

\acro{ROI}{region of interest}
\acrodefplural{ROI}[ROIs]{regions of interest}

\acro{ROS}{Robot Operating System}

\acro{ROV}{remotely operated vehicle}

\acro{RRT}{Rapidly-exploring Random Tree}
\acrodefplural{RRT}[RRTs]{Rapidly-exploring Random Trees}

\acro{SLAM}{Simultaneous Localisation and Mapping}

\acro{SSE}{sum of squared errors}

\acro{STOMP}{Stochastic Trajectory Optimization for Motion Planning}

\acro{TRN}{Terrain-Relative Navigation}

\acro{UAV}{unmanned aerial vehicle}

\acro{USBL}{Ultra-Short Baseline}

\acro{IPP}{informative path planning}

\acro{FoV}{field of view}
\acrodefplural{FoV}[FoVs]{fields of view}

\acro{CDF}{cumulative distribution function}

\acro{ML}{maximum likelihood}

\acro{RMSE}{Root Mean Squared Error}
\acro{MLL}{Mean Log Loss}

\acro{GP}{Gaussian Process}
\acrodefplural{GP}[GPs]{Gaussian Processes}

\acro{KF}{Kalman Filter}

\acro{IP}{Interior Point}
\acro{BO}{Bayesian Optimization}

\acro{SE}{squared exponential}

\acro{UI}{uncertain input}

\acro{MCL}{Monte Carlo Localisation}
\acro{AMCL}{Adaptive Monte Carlo Localisation}

\acro{SSIM}{Structural Similarity Index}

\acro{MAE}{Mean Absolute Error}
\acro{RMSE}{Root Mean Squared Error}
\acro{AUSE}{Area Under the Sparsification Error curve}

\acro{AL}{active learning}
\acro{DL}{deep learning}
\acro{CNN}{convolutional neural network}

\acro{MC}{Monte-Carlo}

\acro{GSD}{ground sample distance}

\acro{BALD}{Bayesian active learning by disagreement}

\acro{fCNN}{fully convolutional neural network}
\acro{FCN}{fully convolutional neural network}

\acro{CMA-ES}{Covariance Matrix Adaptation Evolution Strategy}

\acro{FoV}{field of view}

\acro{mIoU}{mean Intersection-over-Union}
\acro{ECE}{expected calibration error}

\end{acronym}

\IEEEoverridecommandlockouts                           % This command is only needed if
\title{Informative Path Planning for Active Learning \\ in Aerial Semantic Mapping}
\author{Julius R\"{u}ckin, Liren Jin, Federico Magistri, Cyrill Stachniss, Marija Popovi\'{c}
\thanks{This work has been funded by the Deutsche Forschungsgemeinschaft (DFG, German Research Foundation) under Germany's Excellence Strategy, EXC-2070 -- 390732324 (PhenoRob). Authors are with the Cluster of Excellence PhenoRob, Institute of Geodesy and Geoinformation, University of Bonn. Cyrill Stachniss is also with the Lamarr Institute for Machine Learning and Artificial Intelligence, Germany.
Corresponding: \texttt{jrueckin@uni-bonn.de}.}%}
}

\begin{document}

\maketitle

\begin{abstract}
Semantic segmentation of aerial imagery is an important tool for mapping and earth observation.
However, supervised deep learning models for segmentation rely on large amounts of high-quality labelled data, which is labour-intensive and time-consuming to generate.
To address this, we propose a new approach for using unmanned aerial vehicles (UAVs) to autonomously collect useful data for model training.
We exploit a Bayesian approach to estimate model uncertainty in semantic segmentation. During a mission, the semantic predictions and model uncertainty are used as input for terrain mapping.
A key aspect of our pipeline is to link the mapped model uncertainty to a robotic planning objective based on active learning. This enables us to adaptively guide a UAV to gather the most informative terrain images to be labelled by a human for model training.
Our experimental evaluation on real-world data shows the benefit of using our informative planning approach in comparison to static coverage paths in terms of maximising model performance and reducing labelling efforts.

% - Motivation sentence: Semantic segmentation [of remote sensing data] is important because

% - Challenge: However, a key challenge is labelling images to train DL models

% - To address this, we propose a new informative planning framework for active learning in aerial semanic mapping scenarios.

% - [Brief description of our approach, very similar to teaser]

% - Experiments show [repeat the CLAIMS]

\end{abstract}

\section{Introduction} \label{S:intro}
% {vision, motivation, existing work, gap, method, results, menu (optional)}
Semantic segmentation of remote sensing data provides valuable information for monitoring and understanding the environment. \Acp{UAV} are popular for aerial image acquisition in many applications, including urban planning~\citep{Potsdam2018,Lenczner2022,Kemker2018,Tuia2009}, precision agriculture~\citep{Popovic2020,Rodriguez2021,stache2021adaptive}, and wildlife conservation~\citep{Kellenberger2019}, due to their relatively low cost and high operational capabilities~\citep{Osco2021,Kellenberger2019}. In parallel, the advent of deep learning for semantic segmentation with \acp{FCN} has opened new frontiers~\citep{Garcia2017,Romera2018}. These developments have expanded the possibilities for the automated interpretation of aerial imagery towards more complex tasks and larger environments. Classical semantic segmentation approaches rely on supervised learning procedures. They require large amounts of training data that need to be labelled in a pixel-wise fashion, which is labour- and time-consuming.

This paper examines the problem of \ac{AL} for efficient training data acquisition in aerial semantic mapping scenarios. Our goal is to guide a \ac{UAV} to collect highly informative images for training a \ac{FCN} for semantic segmentation in a targeted way, and thus minimise the total amount of labelled data necessary. This is achieved by replanning the UAV paths online as new data is gathered to maximise an existing \ac{FCN}'s segmentation performance when it is retrained on images collected during the mission and labelled by a human operator.

To alleviate the labelling effort required to train \acp{FCN}, \ac{AL} methods have been widely used to optimise the collection of training data. In \ac{AL}, the aim is to selectively label data in order to maximise model performance. Recently, several studies have proposed \ac{AL} frameworks for deep learning models with image data~\citep{Gal2017,Blok2021}, which effectively reduce labelling requirements. However, applying such methods in the context of autonomous robotic monitoring missions has been largely unaddressed. Recent works examining \ac{AL} with \ac{UAV}-acquired imagery~\citep{Kellenberger2019,Lenczner2022} only consider selecting images from a static pool obtained from previous exhaustive aerial surveys. Therefore, linking the \ac{AL} objective to active robotic decision-making and planning remains an open challenge.

\begin{figure}[!t]
    \centering
    \includegraphics[width=\columnwidth]{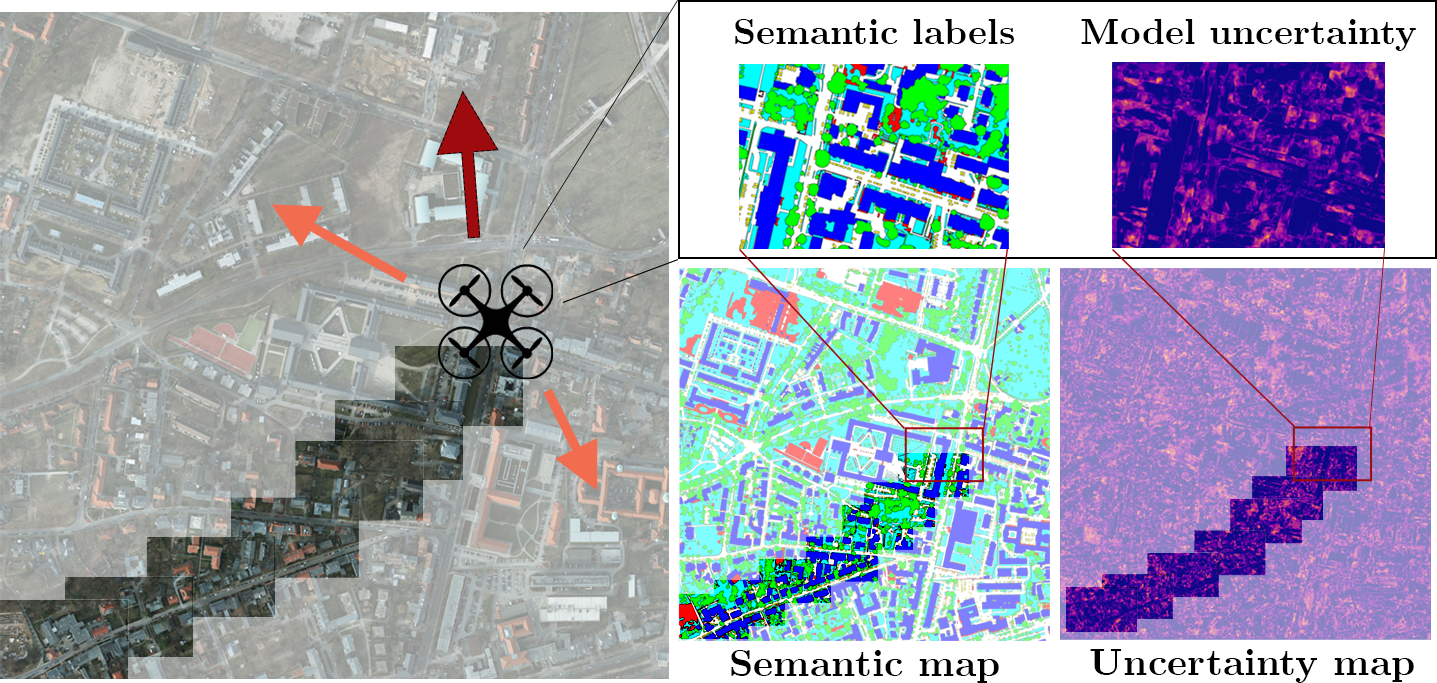}
    \caption{Our planning strategy for active learning in UAV-based terrain mapping. During a mission, we estimate model uncertainty in semantic segmentation (top-right) and fuse it in a global terrain map (bottom-right). Based on the map, our approach guides a UAV to collect most useful (most uncertain) training images for labelling (left). Orange arrows indicate candidate paths and red shows the chosen path.
    In this way, our pipeline reduces the number of images that must be labelled by a human.}
    \label{F:teaser}
\end{figure}

To tackle this issue, we propose a new informative path planning framework for \ac{AL} in aerial semantic mapping tasks, bridging the gap between recent advances in \ac{AL} and robotic applications.  
Our main contribution is a novel approach to integrate \ac{AL} into a planning strategy for efficient training data acquisition.
Our method exploits Bayesian deep learning~\citep{Gal2017} to estimate the model uncertainty of a pretrained \ac{FCN} based on the light-weight ERFNet architecture~\citep{Romera2018}. The \ac{FCN} provides pixel-wise semantic labels and model uncertainties which are fused online into a global terrain map. Based on the model uncertainty, we plan paths for a \ac{UAV} to adaptively collect the most uncertain, i.e. the most informative, images for labelling and retraining the \ac{FCN} after the mission. This pipeline enables reducing labelling efforts without relying on large static pools of unlabelled data. \cref{F:teaser} shows an overview of our approach.

In sum, we make the following three claims.
First, our Bayesian ERFNet achieves higher segmentation performance than its non-Bayesian counterpart and provides consistent model uncertainty estimates for \ac{AL}.
Second, our online planning framework reduces the number of labelled images needed to maximise segmentation performance compared to static coverage paths~\citep{Galceran2013}, which are commonly used in aerial surveys.
Third, we show that map-based planning with a lookahead in our integrated framework yields better \ac{AL} performance compared to greedy one-step planning strategies.
%Third, we show that our proposed planning with lookahead strategy is more beneficial for \ac{AL} performance than greedily optimizing the next-best measurement location.
%(i) An informative path planning framework for \ac{AL} in autonomous robotic data collection scenarios. Our approach enables targeted collection of new training data to maximise the performance of our Bayesian ERFNet, thereby reducing labelling effort requirements.
%(ii) An extension of the ERFNet architecture ~\citep{Romera2018}, Bayesian ERFNet, for probabilistic semantic segmentation on resource-constrained platforms. Bayesian ERFNet improves segmentation performance by $3.8\%$ relative mIoU gain over ERFNet. Further, we validate the improved model uncertainty estimates.
%Our framework will be open-sourced for usage and further development by the community.
We open-source our code for usage by the community.\footnote{\texttt{github.com/dmar-bonn/ipp-al}}

\section{Related Work} \label{S:related_work}

Our work combines recent developments in \ac{AL} and Bayesian deep learning with informative path planning. The aim is to efficiently collect image data for training a deep learning model using autonomous robots.

\subsection{Active Learning with Image Data} \label{SS:active_learning}

\ac{AL} has been intensively researched in the last decades. It aims to maximise a model's performance gain while manually labelling the fewest amount of samples possible from a pool of unlabelled data.
%In computer vision, the goal of \ac{AL} is to select the most informative images from a pool of unlabelled data for training.
Early techniques for \ac{AL} with images focus on kernel-based methods such as support vector machines~\citep{Tuia2009} and Gaussian processes~\citep{Li2013}. Recently, \acp{FCN} have become the state of the art in computer vision due to their superior performance~\citep{Romera2018,Garcia2017,Osco2021}. \ac{AL} has significant potential to reduce human labelling effort for \acp{FCN}, whose performance depends on large high-quality labelled datasets~\citep{Ren2021, Gal2017}. To select informative samples, \ac{AL} methods use acquisition functions to estimate either model uncertainty~\citep{Gal2017} or train data diversity~\citep{Sener2018}. We focus on model uncertainty measures since we aim to detect misclassifications in heterogenous aerial imagery, which may cover large areas.
Building up on ideas by \citet{Gal2017} to extract model uncertainty, we combine concepts from Bayesian deep learning into an \ac{AL} framework for semantic segmentation.

\ac{AL} techniques hold particular promise in remote sensing applications where unlabelled data is abundant but labelled benchmark datasets are scarce. As a result, they are experiencing rapid uptake in various scenarios, e.g. wildlife detection~\citep{Kellenberger2019}, land cover analysis~\citep{Rodriguez2021,Lenczner2022,Blum2019}, and agricultural harvesting~\citep{Blok2021}. \citet{Kellenberger2019} combine \ac{AL} with UAV imagery to improve the performance of an object-detection model subject to domain shifts. However, their \ac{AL} objective is not suitable for quantifying model uncertainty, and assumes large unlabelled data pools from manually executed data campaigns. \citet{Lenczner2022} propose an interactive \ac{AL} framework to obtain accurate aerial segmentation maps. Although they exploit model uncertainty to target misclassified regions, their framework also requires a static annotated database.
%In large-scale monitoring scenarios using satellite imagery, ~\citet{Rodriguez2021} introduce an \ac{AL} approach, which balances model uncertainty and diversity. 
In contrast to these works, we integrate the \ac{AL} objective into a path planning framework. Our approach enables guiding a robot to collect the most informative new training images during a mission, without relying on previously acquired data.
%efficient, targeted data acquisition using resource-constrained robotic systems.

\subsection{Uncertainty in Deep Learning} \label{SS:uncertainty_dl}

Uncertainty quantification is a long-standing problem in machine learning and a key component of \ac{AL}. In our approach, quantifying model uncertainty is crucial as it can be reduced by adding more training samples.
%However, the model posterior distribution, required to quantify uncertainty, is intractable for large models. This makes efficient uncertainty estimation in deep learning particularly challenging~~\citep{Kendall2017uncertainties}.
In deep learning, efficient uncertainty estimation using analytic methods is challenging since the model posterior distribution is often intractable~\citep{Kendall2017uncertainties}.
One approach to estimate model uncertainty is to measure the disagreement between predictions of different networks in an ensemble~\citep{Georgakis2021}. To quantify uncertainty of a single model with reduced computational costs, \ac{MC} dropout is commonly applied~\citep{Gal2016dropout,Kendall2017uncertainties}. At test time, this approach averages multiple model predictions with independently sampled dropout masks. \citet{Kendall2017bayesian} leverage \ac{MC} dropout in a Bayesian \acp{FCN} for semantic segmentation. They show that modelling uncertainty improves segmentation performance and produces a reliable measure for decision-making.
%generalise the \ac{FCN} encoder-decoder architecture to
Similar techniques also have been used to select informative samples in \ac{AL}~\citep{Blok2021,Rodriguez2021, Gal2017}. Our Bayesian ERFNet follows this line of work to obtain computationally efficient model uncertainty calculation for online path replanning.

\subsection{Informative Path Planning for Active Sensing} \label{SS:informative_planning}

In active sensing, informative path planning aims to maximise the information of terrain measurements collected subject to platform constraints, such as mission time. Recent advances have unlocked the potential of efficient data-gathering using autonomous robots in various domains~\citep{Popovic2020,Blum2019,Georgakis2021,Hollinger2013}. Our work considers an adaptive planning strategy which allows a robot to focus on informative training images by replanning online as new data are accumulated~\citep{Hollinger2013}.

Adaptive approaches seek to maximise the expected information gain of new measurements with respect to terrain maps. Such methods are used to, e.g., find hotspots~\citep{Singh2009} or perform inspection~\citep{Hollinger2013} in a targeted way. Several studies consider image processing using \acp{FCN} as a basis for mapping~\citep{Dang2018,Popovic2020, stache2021adaptive}. However, they focus on planning with respect to the semantic predictions. As such, they leverage fully trained models and do not consider model uncertainty.

\begin{figure*}[!t]
    \centering
    \includegraphics[width=0.9\linewidth]{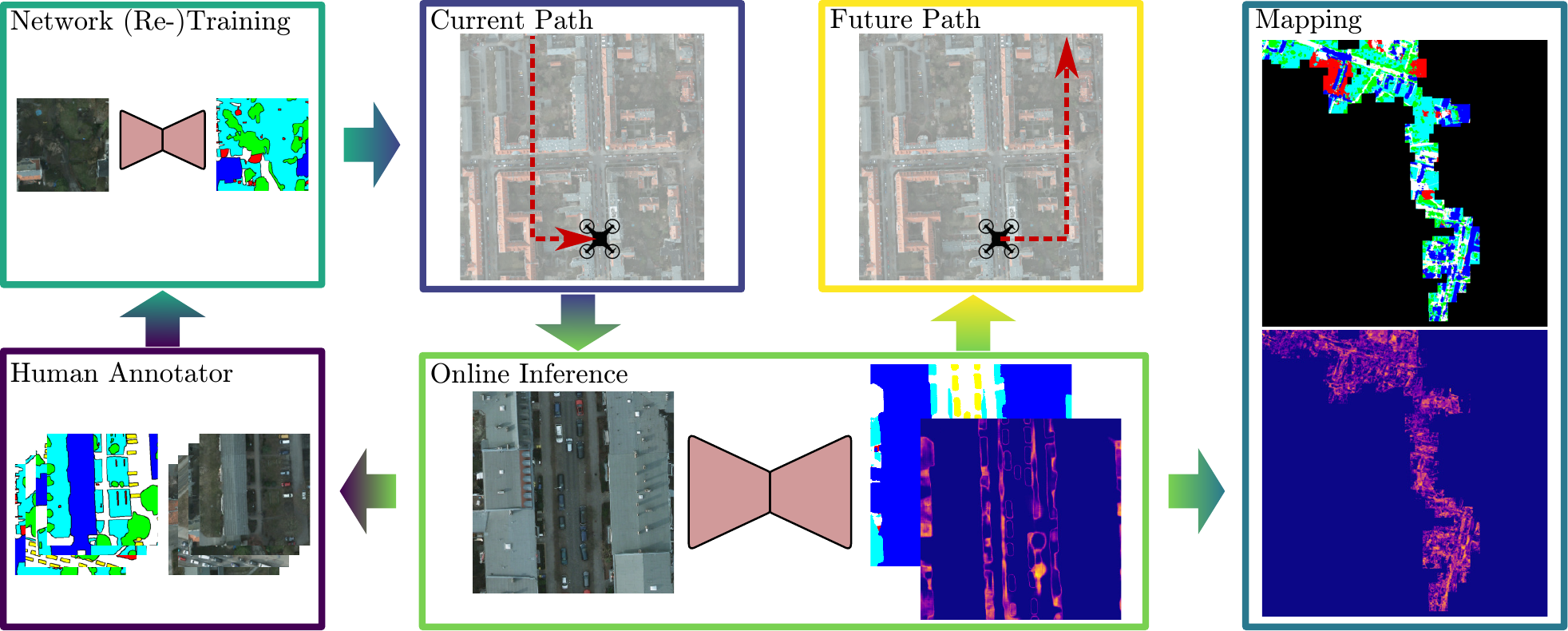}
    \caption{Overview of our proposed approach. We start with a pretrained network for probabilistic semantic segmentation, deployed on a UAV. During a mission, the network processes RGB images to predict pixel-wise semantic labels and model uncertainties (\cref{SS:network_architecture}), which are projected onto the ground to build global maps capturing these variables (\cref{SS:mapping}). Based on the estimated model uncertainty, the current UAV position, and the current map state, our algorithm plans paths for the UAV to collect the most uncertain (most informative) training data for improving the network performance (\cref{SS:path_planning}). After the mission, the collected images are labelled by an annotator and used for network retraining. By guiding the UAV to collect informative training data, our pipeline reduces the labelling effort.}
    \label{F:approach_overview}
\end{figure*}

The combination of path planning with \ac{AL} is a relatively unexplored research area, but presents a natural way to exploit current robotic capabilities for reducing labelling efforts. \citet{Georgakis2021} propose a framework for active semantic goal navigation which uses an ensemble to estimate model uncertainty as the planning objective. In contrast, our Bayesian ERFNet exploits \ac{MC} dropout and a light-weight architecture for fast online replanning in \ac{UAV}-based scenarios. Most resembling our work is the planning approach of \citet{Blum2019} for \ac{AL} in semantic mapping. A key difference to our approach is that their algorithm relies on novelty, which requires reasoning about the entire growing training set. In contrast, our \ac{MC} dropout procedure permits fast model uncertainty estimation suitable for replanning. Moreover, we leverage model uncertainties to maintain a consistent terrain map as a basis for global planning.

% Do we need this one?
% Decomposition of aleatoric and model uncertainty in Bayesian DL for active learning and risk-aware RL in the presence of heteroscedastic noise ~\cite{depeweg2018decomposition}.

\section{Our Approach} \label{S:approach}
We propose a new informative path planning framework for \ac{AL} in \ac{UAV}-based semantic mapping scenarios. Our goal is to deploy a \ac{UAV} to adaptively collect the most informative training images to be labelled by a human, and thereby reduce labelling efforts. \cref{F:approach_overview} depicts an overview of our framework. During a mission, we use a pretrained \ac{FCN} to perform semantic segmentation on aerial images of terrain and estimate the associated model uncertainty. The model prediction uncertainty are then used as input for semantic terrain mapping. In this setup, we exploit model uncertainty estimates to plan paths online for gathering new data that maximises an \ac{AL} information objective.
The following subsections describe the key elements in our proposed approach. Note that these modules are generic and can be adapted for any given monitoring scenario. Thus, we see our main contribution as a general robotic planning framework for \ac{AL} that enables efficient training data acquisition.

\subsection{Bayesian Segmentation Network} \label{SS:network_architecture}

\begin{figure}[!t]
    \centering
    \includegraphics[width=0.84\linewidth]{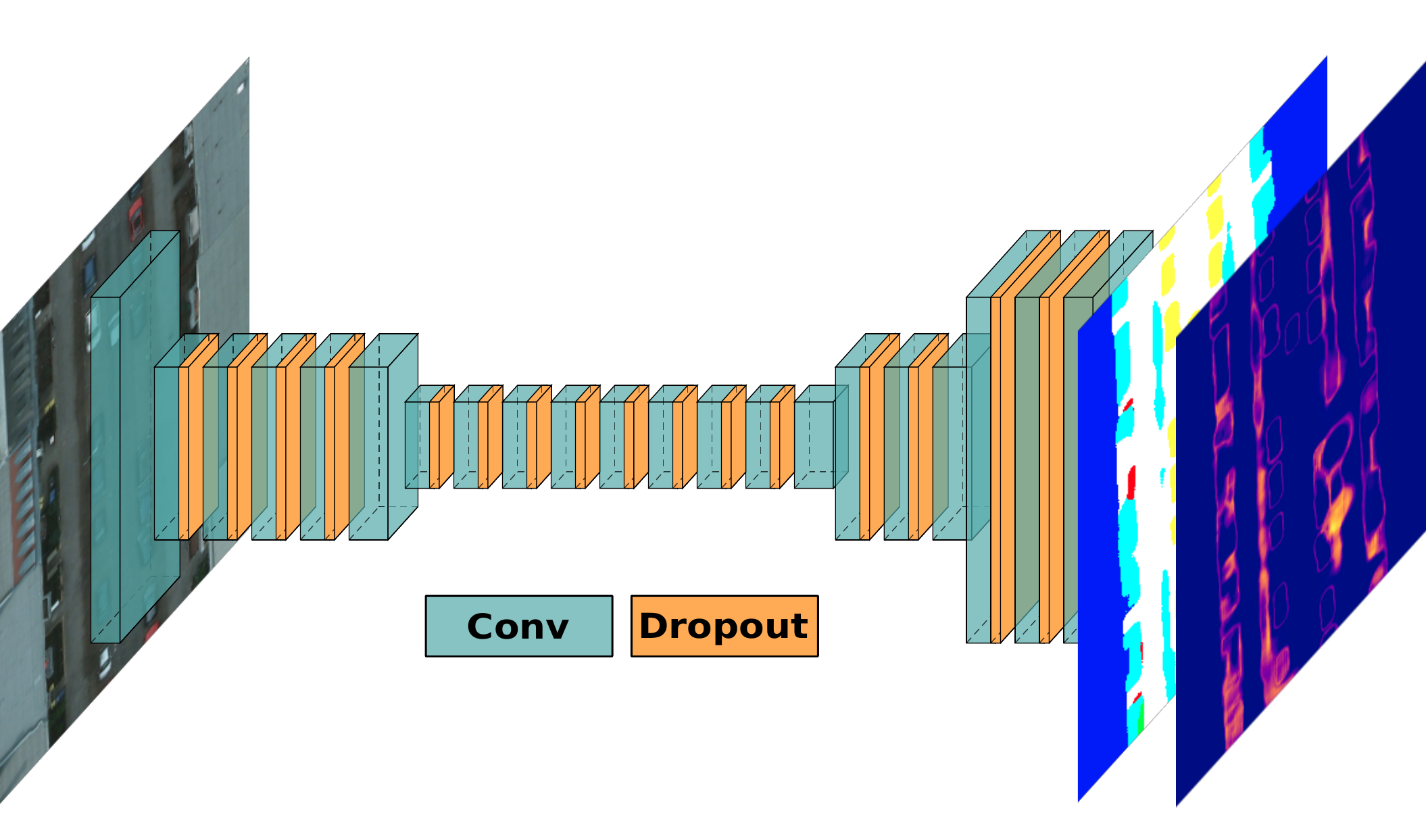}
    \caption{Our Bayesian ERFNet architecture for probabilistic semantic segmentation. We extend the network of \citet{Romera2018} with Monte-Carlo dropout (orange layers) to predict model uncertainty. Our network takes as input RGB (left) and outputs semantic labels (second from right) and pixel-wise uncertainty (first from right).}
    \label{F:network_architecture}
\end{figure}

Our goal is to perform the semantic segmentation on RGB images and predict the associated pixel-wise model uncertainty as a basis for \ac{AL}. To achieve this, we extend the ERFNet architecture proposed by \citet{Romera2018} to account for this probabilistic model interpretation. \cref{F:network_architecture} depicts our new network.
Our main motivation for using ERFNet is its effectiveness for real-time inference, as required for online replanning, which is essential in our setup.
 
ERFNet is an encoder-decoder \ac{FCN} structure with 16 blocks in the encoder and 5 blocks in the decoder. A key feature is the use of non-bottleneck-1D blocks, separating 2D convolutional layers into two subsequent 1D convolutional layers to reduce the learnable parameter count without sacrificing model capacity. The model $\bm{f}^{\bm{W}}(\bm{z})$ is parameterised by weights $\bm{W}$ and its output is the likelihood $p(\bm{y} \mid \bm{f}^{\bm{W}}(\bm{z})) = \text{softmax}(\bm{f}^{\bm{W}}(\bm{z}))$, where $\bm{z}$ is the input RGB image, and $\bm{y}$ is the predicted pixel-wise class probability simplex. 

We leverage advances in Bayesian deep learning to create a probabilistic version of ERFNet quantifying model uncertainty for \ac{AL}~\citep{Gal2016dropout, Kendall2017uncertainties}. Assume a train set with images $\bm{Z} = \{\bm{z}_1, \bm{z}_2, ...\}$ and labels $\bm{Y} = \{\bm{y}_1, \bm{y}_2, ...\}$. To quantify the model uncertainty, we estimate the posterior $p(\bm{W}\,|\,\bm{Z}, \bm{Y})$ with \ac{MC} dropout~\citep{Gal2016dropout} performing multiple forward passes at test time with independently sampled dropout masks.
%, which can be seen as performing variational inference with a simple distribution $q^{*}_{\theta}(\bm{W})$ over the network weights, where $q^{*}_{\theta}$ is parameterised by $\theta$. 
Thus, we add dropout layers after all non-bottleneck-1D blocks in the encoder and decoder to create a fully Bayesian ERFNet. Our network is trained to minimise cross entropy~\citep{Kendall2017uncertainties}:
\begin{equation} \label{eq:loss}
    \mathcal{L}(\theta) = - \frac{1}{\,|\,\bm{Z}\,|\,} \sum_{i = 1}^{\,|\,\bm{Z}\,|\,} \text{log} \, p(\bm{y}_i \,|\,\bm{f}^{\hat{\bm{W}_i}}(\bm{z}_i)) + \lambda \lVert \bm{W} \rVert_{2}^{2},
\end{equation}
where $\lambda$ is the weight decay and weights $\hat{\bm{W}_i}$ are sampled by applying dropout to the full set of parameters $\bm{W}$.

At deploy time, the prediction $\bm{y}$ for an image $\bm{z}$ is computed by marginalising over the approximated model posterior. To do so in a tractable fashion, we use \ac{MC} integration~\citep{Kendall2017uncertainties}:
\begin{equation} \label{eq:predictive_dist}
    \hat{p}(\bm{y}\,|\,\bm{z}, \bm{Z}, \bm{Y}) = \frac{1}{T} \sum_{t=1}^{T} \text{softmax}(\bm{f}^{\hat{\bm{W}}_t}(\bm{z})),
\end{equation}
where $\hat{\bm{W}_t}$ are independently sampled by applying dropout to the full set of parameters $\bm{W}$ in each forward pass.
%where $\hat{\bm{W}}_t \sim q^{*}_{\theta}(\bm{W})$ is equivalent to independently sampling dropout masks in each forward pass $t \in [T]$. 

Next, we outline our approach for linking the model uncertainty to an \ac{AL} acquisition function. This function encodes the expected value of adding a new image to the training dataset and is used to select most informative samples for annotation in \cref{F:approach_overview}. The main idea is to pick images with the highest model uncertainty, i.e. highest information value, for training.
To measure information, we use the Bayesian active learning by disagreement acquisition function~\citep{houlsby2011bayesian}, where images from an unlabelled pool are chosen to maximise the mutual information $\mathbb{I}(\bm{y}, \bm{W}\,|\,\bm{z}, \bm{Z}, \bm{Y}) \in [0, 1]$ between model predictions and the posterior. Intuitively, this quantity is high when the predictions induced by the sampled network weights differ substantially, while the model is certain in each single prediction. 
%To estimate model uncertainty of the posterior prediction $\hat{p}(\bm{y}\,|\,\bm{z}, \bm{Z}, \bm{Y})$, we leverage the mutual information $\mathbb{I}(\bm{y}, \bm{W}\,|\,\bm{z}, \bm{Z}, \bm{Y}) \in [0, 1]$ between model predictions and model posterior. Intuitively, this quantity is high when the predictions induced by the sampled network weights differ substantially while the model is certain in each single prediction. At the same time, $\mathbb{I}(\bm{y}, \bm{W}\,|\,\bm{z}, \bm{Z}, \bm{Y})$ defines the \ac{BALD} acquisition function for \ac{AL} ~\citep{houlsby2011bayesian}, where images from an unlabelled pool are chosen to maximise $\mathbb{I}(\bm{y}, \bm{W}\,|\,\bm{z}, \bm{Z}, \bm{Y})$.
%Thus, it is particularly suitable for our use case as the model uncertainty directly links to the \ac{AL} objective. 
However, for our Bayesian ERFNet, the model posterior prediction $\hat{p}(\bm{y}\,|\,\bm{z}, \bm{Z}, \bm{Y})$ is intractable. This prevents us from computing model uncertainty, and hence information gain, analytically. Instead, following \citet{Gal2017}, we approximate mutual information by using \cref{eq:predictive_dist}:
\begin{multline}  \label{eq:mutual_information}
    \mathbb{I}(\bm{y}, \bm{W}\,|\,\bm{z}, \bm{Z}, \bm{Y}) ~ \approx ~ -\hat{p}(\bm{y}\,|\,\bm{z}, \bm{Z}, \bm{Y})^{T}\text{log}\big(\hat{p}(\bm{y}\,|\,\bm{z}, \bm{Z}, \bm{Y})\big) \\
    ~+~ \frac{1}{T} \sum_{t=1}^{T} ~ p(\bm{y}\,|\,\bm{z}, \hat{\bm{W}}_t)^{T}\text{log}\big(p(\bm{y}\,|\,\bm{z}, \hat{\bm{W}}_t)\big),
\end{multline}
where $\text{log}(\cdot)$ is applied element-wise. For the derivation, we refer to the work of \citet{Gal2017}.

\subsection{Terrain Mapping} \label{SS:mapping}

A key aspect of our approach is a probabilistic map, which captures the semantic and uncertainty information about the 2D terrain at a given time. This map is updated online during a mission as new measurements arrive, and can be exploited for global uncertainty-based planning.
Our method leverages sequential Bayesian fusion and probabilistic occupancy grid mapping~\citep{Elfes1989} to map semantic terrain information and model uncertainty respectively. The terrain $\xi \subset \mathbb{R}^2$ is discretised using two grid maps: a semantic map $\mathcal{X}_{s}$ and an model uncertainty map $\mathcal{X}_{u}$. The semantic information is represented by $\zeta: \mathcal{X}_{s} \to [0,1]^C$ and we assign a $C$-dimensional probability simplex to each discrete grid cell on the terrain $\xi$, where $C$ is the number of possible class labels. Similarly, the model uncertainty in each cell is given by $\eta: \mathcal{X}_{u} \to [0,1]$. 

To map multiple semantic classes, we maintain one independent grid layer in $\mathcal{X}_{s}$ for each of the $C$ labels. In each layer $i \in [C]$, the prior semantic map distribution $p(\zeta_{i}\,|\,\xi) \sim \mathcal{N}(\bm{\mu^{-}},\, \bm{P^{-}})$, is defined by a mean vector $\bm{\mu^{-}} = \frac{1}{C} \mathbf{1}$ and a diagonal covariance matrix $\bm{P^{-}} = \varepsilon \bm{I}$ with variances $\varepsilon \in (0, 1]$. During a mission, when an image is taken at time step $t$, the resulting output from our network is projected to the flat terrain given the camera intrinsics and \ac{UAV} position. To update the map, a Kalman filter is used to recursively fuse the pixel-wise probabilistic segmentation predictions $\bm{s}_{t} = \hat{p}(\bm{y}\,|\,\bm{z}_{t}, \bm{Z}, \bm{Y})$ (\Cref{eq:predictive_dist}) and estimated model uncertainties $\bm{u}_{t} = \mathbb{I}(\bm{y}, \bm{W}\,|\,\bm{z}_{t}, \bm{Z}, \bm{Y})$ (\Cref{eq:mutual_information}) based on input image $\bm{z}_{t}$ taken at measurement position $\bm{x}_{t} \in \mathbb{R}^3$ above the terrain $\xi$ into a posterior mean $\bm{\mu^{+}}$ and a posterior diagonal covariance matrix $\bm{P^{+}}$ of model uncertainties defining the posterior map state $\mathcal{X}_{s, t} \sim \mathcal{N}(\bm{\mu^{+}}, \bm{P^{+}})$ at time step $t$. Note that we apply standard Bayesian fusion to update the map state $\mathcal{X}_{s, t}$.

To handle model uncertainties in $\mathcal{X}_u$, our goal is to fuse new uncertainties at time step $t$ into the map as well as to keep track of exploration frontiers for planning. To achieve this, we update: (1) a maximum likelihood map state $\mathcal{X}_{u, t}$ of model uncertainty estimates $\bm{u}_{t}$ at measurement positions $\bm{x}_t$ and (2) a hit map $\mathcal{H}_{u, t}$ counting total updates of grid cells in $\mathcal{X}_{u, t}$, i.e. hits. By combining standard robotic mapping techniques with Bayesian deep learning, we maintain a terrain map suitable for global planning with an \ac{AL} objective. We note the differences between our approach and previous works~\citep{Blum2019,Georgakis2021}, which assess the expected information gain of new data for \ac{AL} only on a local per-image basis, without considering its global distribution in the environment.

\subsection{Informative Path Planning} \label{SS:path_planning}

We develop informative path planning algorithms to guide a UAV to adaptively collect useful training data for our \ac{FCN}. Our goal is to plan paths to gather new images that maximise the model uncertainty, i.e. maximise the \ac{AL} acquisition function, as described in \cref{SS:network_architecture}.

In general, the informative path planning problem aims to maximise the information gain $\mathrm{I}(\cdot)$ of measurements collected within an initially unknown environment $\xi$:
\begin{equation} \label{eq:ipp_objective}
    \psi^{*} = \argmax_{\psi \in \Psi}\, \mathrm{I}(\psi),\, \text{s.t. } \mathrm{C}(\psi) \leq \mathrm{B},
\end{equation}
where $\mathrm{C}: \Psi \to \mathbb{R}_{\geq 0}$ maps a path $\psi$ to its execution costs, $\mathrm{B} \in \mathbb{R}_{\geq 0}$ is the robot's budget limit, e.g. time or energy, and $\mathrm{I}: \Psi \to \mathbb{R}_{\geq 0}$ is the information criterion, computed from the new measurements taken along $\psi$. In our work, the costs $\mathrm{C}(\psi)$ of a path $\psi = (\bm{x}_1, \ldots, \bm{x}_N)$ of length $N$ are defined by the total flight time:
\begin{equation} \label{eq:ipp_cost_fn}
    \mathrm{C}(\psi) = \sum_{i=1}^{N-1} \mathrm{c}(\bm{x}_i, \bm{x}_{i+1}),
\end{equation}
where $\bm{x}_{i} \in \mathbb{R}^3$ is a 3D position above the terrain $\xi$ an image is registered from. The function $\mathrm{c}: \mathbb{R}^3 \times \mathbb{R}^3 \to \mathbb{R}_{\geq 0}$ computes the flight time between measurement positions by a constant acceleration-deceleration $\pm u_{a}$ with maximum speed $u_{v}$. A key feature of our approach is to relate the information criterion $\mathrm{I}(\psi)$ to the \ac{AL} acquisition function, i.e. we plan paths to collect new images along $\psi$ that maximise the model's performance gain when added to the training set.
%the value of measurements along $\psi$ for \ac{AL}, i.e. maximising the model's performance gain by extending its training data with the newly collected images along $\psi$.

\begin{figure}[!t]
    \centering
    \begin{subfigure}{0.32\columnwidth}
        \includegraphics[width=\columnwidth]{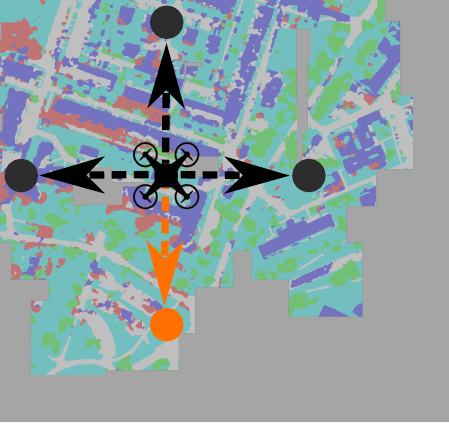}
        \caption{Image-based} \label{SF:image}
    \end{subfigure}
    \begin{subfigure}{0.32\columnwidth}
        \includegraphics[width=\columnwidth]{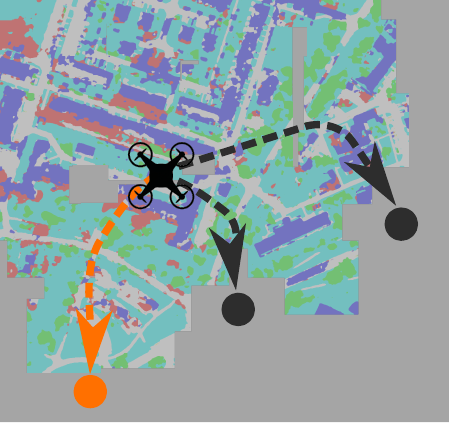}
        \caption{Frontier-based} \label{SF:frontier}
    \end{subfigure}
    \begin{subfigure}{0.32\columnwidth}
        \includegraphics[width=\columnwidth]{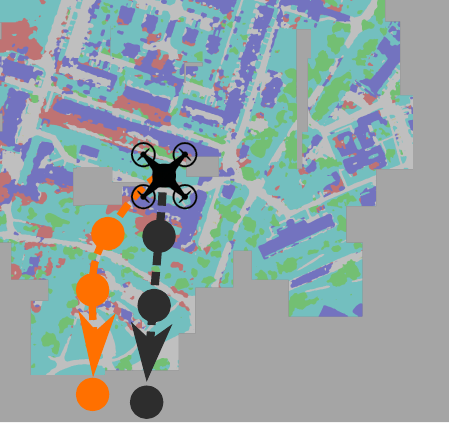}
        \caption{Fixed-horizon} \label{SF:fixed}
    \end{subfigure}
    \caption{Our planning strategies for training data acquisition. Black dots and arrows indicate candidate measurements and paths evaluated based on their information gain adding new images to the training set. Orange dots and arrows indicate the best chosen measurement and paths, respectively. Grey depicts unexplored areas.}
    \label{F:planning_strategies}
\end{figure}

Next, we propose three different informative planning strategies to optimise \cref{eq:ipp_objective} for replanning during a mission. The approaches are illustrated in \cref{F:planning_strategies}. In our experiments in \cref{SS:planning_evaluation}, we compare each planner in terms of segmentation performance over the total labelling cost.
%and show that our proposed planning with lookahead strategy is more beneficial for \ac{AL} performance than greedily optimising the next-best measurement location.

\noindent \textbf{Image-based planner}. Our image-based planner follows the direction of the highest estimated model uncertainty in the image recorded at the current UAV position (\cref{SF:image}). To this end, we greedily plan a next-best measurement position $\bm{x}^{*}_{t}$ at time step $t$ based on the previously estimated model uncertainty $\bm{u}_{t-1} \in [0, 1]^{W \times H}$ over segmentation $\bm{s}_{t-1} \in [0, 1]^{C \times W \times H}$ of RGB image $\bm{z}_{t-1}$ of size $W \times H$, and the current hit map state $\mathcal{H}_{u, t}$. The main idea is to choose the next-best measurement position $\bm{x}^{*}_{t}$ towards the image edge  $e^{*}_{t}$ maximising the model uncertainty over hit counts in $\mathcal{H}_{u, t}$. To do so, we set a width $\leq \min(\frac{W}{2},\frac{H}{2})$ pixels defining the edge area, and sum pixel-wise model uncertainty $\bm{u}_{t-1}$ along each edge normalised over the respective sum of grid cell counts in $\mathcal{H}_{u, t}$. Finally, $\bm{x}^{*}_{t}$ is computed by moving from the previous position $\bm{x}_{t-1}$ in direction of $e^{*}_{t}$ for a user-defined step size at a fixed altitude. Note that this strategy relies only on model uncertainty in the current local image and does not exploit our global uncertainty map. This resembles the planner proposed by \citet{Blum2019}.

\noindent \textbf{Frontier-based planner}. Our frontier-based planner navigates to the most uncertain boundaries of explored terrain, given the current state of the global uncertainty map (\cref{SF:frontier}). We greedily select a next-best measurement point $\bm{x}^{*}_{t}$ at the boundary of known and unknown space of the terrain $\xi$ at time step $t$. 
%The frontiers are determined by finding the contours in the current binary hit map state $\mathcal{H}^{binary}_{u, t}[i,j] = 1$, if $\mathcal{H}^{binary}_{u, t}[i,j] > 0$ for grid cell $[i,j]$. 
Candidate frontiers are generated by sampling positions equidistantly along the frontiers at a fixed altitude. Then, $\bm{x}^{*}_{t}$ is chosen to be the frontier position maximising the model uncertainty in the global map $\mathcal{X}_{u,t}$ normalised over hit counts in $\mathcal{H}_{u, t}$, corresponding to the camera \ac{FoV} at the respective frontier position.
%
%\begin{equation} \label{eq:frontier_objective}
%    \bm{x}^{*}_{t} = \argmax_{\bm{x}_{f, t} \in \mathcal{F}_{t}} \frac{\lVert %\mathcal{X}_{u,t}^{sub}(\bm{x}_{f, t}) \rVert_1}{\lVert \mathcal{H}^{sub}_{u, t}(\bm{x}_{f, t}) %\rVert_1},
%\end{equation}
%
% where $\mathcal{X}_{u,t}^{sub}(\bm{x}_{f, t})$ and $\mathcal{H}^{sub}_{u, t}(\bm{x}_{f, t})$ are the model uncertainty and the hit map's grid subsets spanned by a candidate image taken at pose $\bm{x}_{f, t}$, respectively.

\noindent \textbf{Fixed-horizon planner}. In contrast to the  previous two strategies, our fixed-horizon planner reasons about images taken over multiple time-steps, given the current global uncertainty map state (\cref{SF:fixed}). To this end, we propose a fixed-horizon strategy inspired by the informative path planning framework for terrain monitoring introduced by \citet{Popovic2020}. To ensure efficient online replanning, we use a two-step approach. First, we greedily optimise a path $\psi^{g}_{t} = (\bm{x}_{t}^{g}, ..., \bm{x}_{t+N-1}^{g})$ of length $N$ at time step $t$ using grid search at a fixed altitude. Second, an optimisation routine is applied to refine $\psi^{g}_{t}$ in the continuous UAV workspace and obtain the next-best path $\psi^{*}_{t} = (\bm{x}_{t}^{*}, ..., \bm{x}_{t+N-1}^{*})$.

In the first step, $\psi^{g}_{t}$ is computed sequentially for $N$ steps by choosing the $n$-th measurement point $\bm{x}_{t + n}^{g}$, $n \leq N - 1$, over a sparse grid of candidate positions $\bm{x}_{c}$:
\begin{equation} \label{eq:greedy_objective}
    \bm{x}_{t + n}^{g} = \argmax_{\bm{x}_{c}} \frac{\lVert \mathcal{X}_{u,t}(\bm{x}_{c}) \rVert_1}{\lVert \mathcal{H}_{u, t+n}(\bm{x}_{c}) \rVert_1 \cdot \mathrm{c}(\bm{x}_{t+n-1}^{g}, \bm{x}_{c})},
\end{equation}
where $\lVert \cdot \rVert_1$ is the matrix norm summing all its elements, and $\mathcal{X}_{u,t}(\bm{x}_{c})$ and $\mathcal{H}_{u, t+n}(\bm{x}_{c})$ are the model uncertainty map's subset at time step $t$ and forward-simulated hit map's subset at future time step $t+n$  spanned by the camera \ac{FoV} at position $\bm{x}_{c}$, respectively.

In the second step, we apply an optimisation routine initialised with $\psi^{g}_{t}$ using an objective function similar to \cref{eq:greedy_objective}, but generalised to candidate paths $\psi_{t}^{c} = (\bm{x}_{t}^{c}, ..., \bm{x}_{t+N-1}^{c})$:
\begin{equation} \label{eq:cmaes_objective}
    \psi_{t}^{c} = \frac{\sum_{n=0}^{N-1} \lVert \mathcal{X}_{u,t}(\bm{x}_{t+n}^{c}) \rVert_1}{\sum_{n=0}^{N-1} \lVert \mathcal{H}_{u, t+n}(\bm{x}_{t+n}^{c}) \rVert_1 \cdot \mathrm{c}(\bm{x}_{t+n-1}^{c}, \bm{x}_{t+n}^{c})},
\end{equation}
where $\bm{x}_{t-1}^{c} = \bm{x}_{t-1}$ is the previous measurement position. After convergence, the best-performing candidate path $\psi_{t}^{*} = (\bm{x}_{t}^{*}, ..., \bm{x}_{t+N-1}^{*})$ with respect to \cref{eq:cmaes_objective} is returned, and $\psi_{t, 0}^{*} = \bm{x}_{t}^{*}$ is executed as the next-best measurement position.

Note that we normalise by the flight time to efficiently use the budget $B$, forward-simulate $\mathcal{H}_{u, t}$ to penalise inspecting the same terrain regions again, and assume a uniform prior in unknown regions of $\mathcal{X}_{u,t}$ to foster exploration.

\section{Experimental Results} \label{S:results}

%This section presents our experimental results. We first validate our Bayesian ERFNet design and model uncertainty quantification method in an ablation study as a basis for \ac{AL}. Then, we describe the \ac{AL} training procedure and assess \ac{IPP} performance for \ac{AL} in aerial semantic mapping scenarios.
Our experimental evaluation aims to analyse the performance of our approach and support the claims that our Bayesian \ac{FCN} achieves higher segmentation performance than its non-Bayesian counterpart and provides consistent model uncertainty estimates for \ac{AL} (\cref{SS:ablation_study}). Further, the experiments will show that our online planning framework reduces the number of labelled images needed to maximise segmentation performance compared to static coverage paths (\cref{SS:planning_evaluation}), and our proposed planning with lookahead strategy is more beneficial for \ac{AL} performance than greedily optimising for the next-best measurement (\cref{SS:planning_evaluation}).

\begin{table*}[!h]
\centering
\begin{tabular}{lccc}
    Variant (dropout probabilities) & {Accuracy [\%] $\uparrow$} & {mIoU [\%] $\uparrow$} & {ECE [\%] $\downarrow$} \\ \midrule[1.2pt]
    Non-Bayesian (10\% / 30\% / 50\%)  & 82.28 / 81.99 / 82.71 &  64.47 / 64.35 / 66.24  &  59.98 / 59.09 / 59.43 \\ \midrule
    Standard (10\% / 30\% / 50\%)  & 82.47 / 82.21 / 83.94 &  64.81 / 64.70 / 68.00  &  57.32 / 55.46 / 55.41 \\ \midrule
    Center (10\% / 30\% / 50\%) & 81.58 / 82.26 / 82.91 & 62.34 / 64.47 / 65.78 & 60.05 / 57.61 / 58.55 \\ \midrule
    Classifier (10\% / 30\% / 50\%) & 80.80 / 82.14 / 81.04 & 62.02 / 63.94 / 62.16  & 60.62 / 60.40 / 61.02 \\ \midrule
    All (10\% / 30\% / 50\%, \cref{SS:network_architecture}) & 84.00 / 82.20 /  $\mathbf{84.24}$ & 67.75 / 63.93 / $\mathbf{68.74}$ & 55.76 / $\mathbf{53.26}$ / 54.87
\end{tabular}
\caption{Ablation study of our Bayesian ERFNet trained with varying dropout layers and probabilities $p = \{10\%,\,30\%,\,50\%\}$ on the ISPRS Potsdam dataset~\citep{Potsdam2018} with $T = 50$ MC dropout samples. Our best Bayesian ERFNet (all, $p = 50 \%$) outperforms non-Bayesian ERFNet (standard, $p = 50 \%$) in terms of segmentation performance (by $3.8 \%$ mIoU) and model uncertainty calibration (by $8.3 \%$ ECE).} \label{T:results_ablation_model}
\end{table*}

\begin{figure}[!t]
    \centering
    \includegraphics[width=0.9\columnwidth]{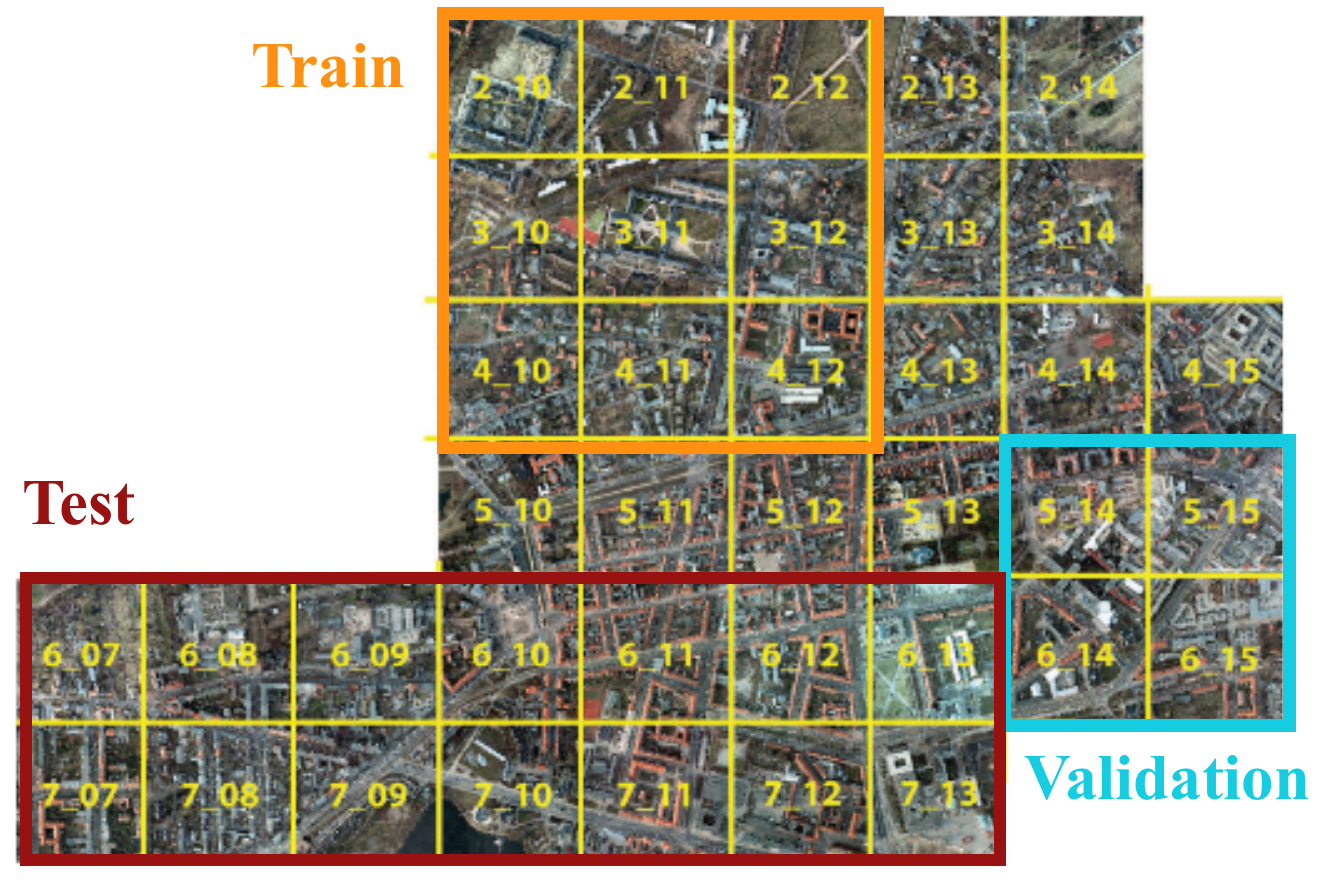}
    \caption{The urban ISPRS Potsdam orthomosaic~\citep{Potsdam2018}. We simulate images and labels with a square footprint, downwards-facing camera, and GSD of $15$\,cm/px to generate training (orange), validation (blue), and testing (red) data from disjoint areas.}
    \label{F:potsdam_dataset}
\end{figure}

\subsection{Bayesian ERFNet Ablation Study} \label{SS:ablation_study}

The experimental results presented in this section aim to support the claim that our Bayesian ERFNet proposed in \cref{SS:network_architecture} achieves higher segmentation performance than the non-Bayesian ERFNet and provides consistent model uncertainty estimates for \ac{AL}.
%Our first aim is to evaluate our Bayesian ERFNet model design and validate that it produces consistent model uncertainty estimates as a basis for \ac{AL}. 
To this end, we perform an ablation study with varying Bayesian units of the ERFNet base architecture to find the best-performing trade-off between segmentation performance and model uncertainty estimation. We assess four different probabilistic variants of ERFNet with varying dropout probabilities $p = \{10\%,\,30\%,\,50\%\}$:
\begin{itemize}
    \item \textbf{Standard}: Dropout layers after all non-bt-1D layers in the encoder as in the normal ERFNet implementation.
    \item \textbf{Center}: Dropout layers after the last four and first two encoder and decoder non-bt-1D layers respectively. 
    \item \textbf{Classifier}: Single dropout layer after the last decoder non-bt-1D layer before the classification head.
    \item \textbf{All}: Dropout layers after all non-bt-1D layers in the encoder and decoder as proposed in \cref{SS:network_architecture}.
\end{itemize}

All models are trained and evaluated on the 7-class urban aerial ISPRS Potsdam orthomosaic dataset~\citep{Potsdam2018}. We simulate images and labels at random uniformly chosen positions from $30$\,m altitude with a square footprint, downwards-facing camera, and \ac{GSD} of $15$\,cm/px. In total, we create 4000, 1000, and 3500 training, validation, and testing datapoints, respectively, from disjoint areas as depicted in \cref{F:potsdam_dataset}. We assess segmentation performance by pixel-wise global accuracy and \ac{mIoU}~\citep{everingham2010pascal}, and model uncertainty estimate quality by the \ac{ECE}. At test time, we use a reasonably large number of \ac{MC} dropout samples $T = 50$.

\begin{figure}[!h]
    \centering
    % row 0
    \begin{subfigure}{0.23\columnwidth}
        \centering
        \includegraphics[width=\columnwidth]{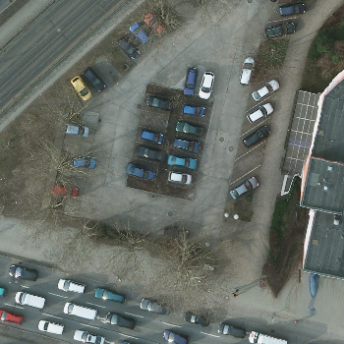}
    \end{subfigure}
    \begin{subfigure}{0.23\columnwidth}
        \centering
        \includegraphics[width=\columnwidth]{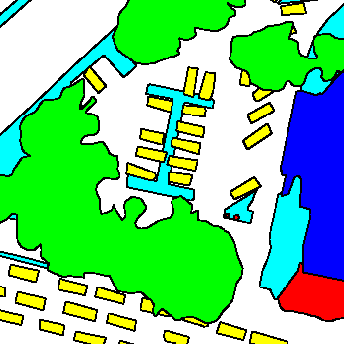}
    \end{subfigure}
    \begin{subfigure}{0.23\columnwidth}
        \centering
        \includegraphics[width=\columnwidth]{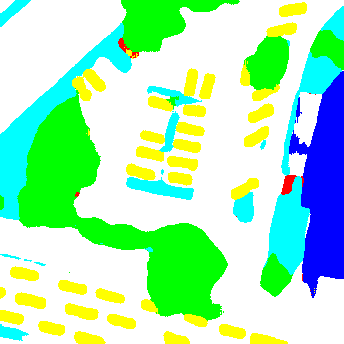}
    \end{subfigure}
    \begin{subfigure}{0.23\columnwidth}
        \centering
        \includegraphics[width=\columnwidth]{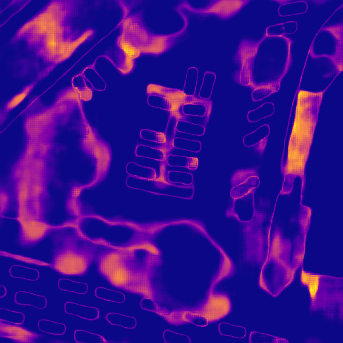}
    \end{subfigure}

    % row 1
    \vspace{1mm}
    \begin{subfigure}{0.23\columnwidth}
        \centering
        \includegraphics[width=\columnwidth]{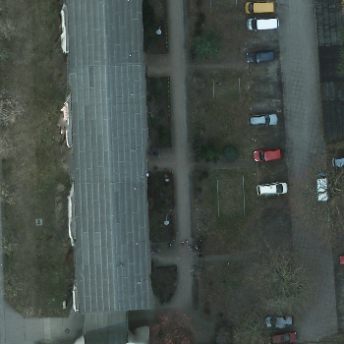}
    \end{subfigure}
    \begin{subfigure}{0.23\columnwidth}
        \centering
        \includegraphics[width=\columnwidth]{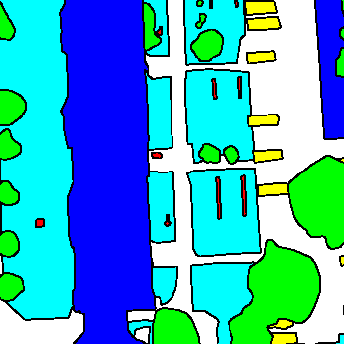}
    \end{subfigure}
    \begin{subfigure}{0.23\columnwidth}
        \centering
        \includegraphics[width=\columnwidth]{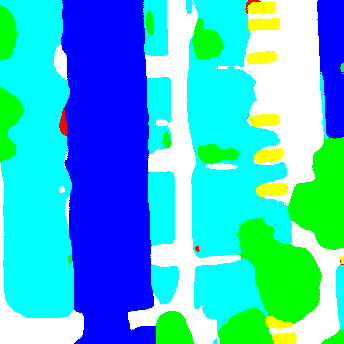}
    \end{subfigure}
    \begin{subfigure}{0.23\columnwidth}
        \centering
        \includegraphics[width=\columnwidth]{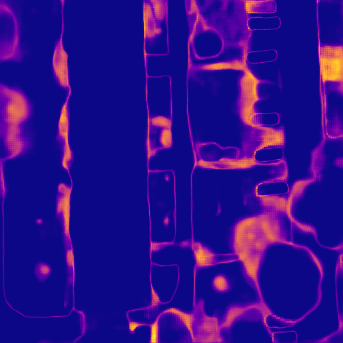}
    \end{subfigure}

    % row 2
    \vspace{1mm}
    \begin{subfigure}{0.23\columnwidth}
        \centering
        \includegraphics[width=\columnwidth]{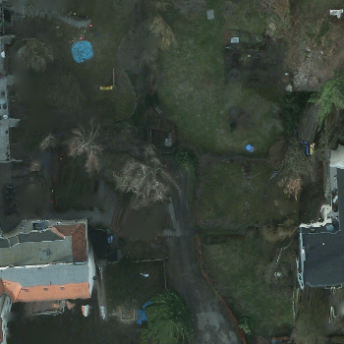}
    \end{subfigure}
    \begin{subfigure}{0.23\columnwidth}
        \centering
        \includegraphics[width=\columnwidth]{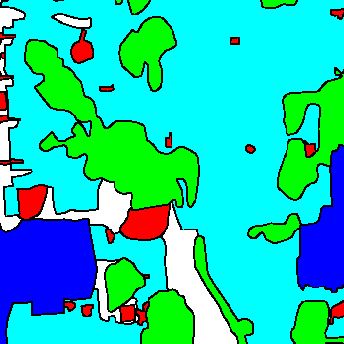}
    \end{subfigure}
    \begin{subfigure}{0.23\columnwidth}
        \centering
        \includegraphics[width=\columnwidth]{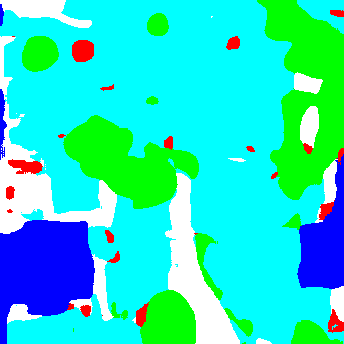}
    \end{subfigure}
    \begin{subfigure}{0.23\columnwidth}
        \centering
        \includegraphics[width=\columnwidth]{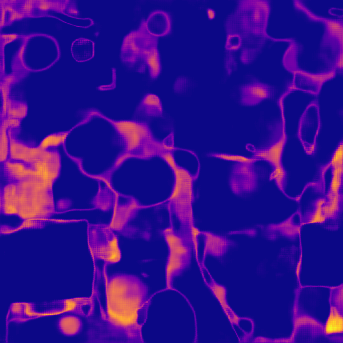}
    \end{subfigure}

    % row 3
    % \vspace{1mm}
    % \begin{subfigure}{0.24\columnwidth}
    %     \centering
    %     \includegraphics[width=\columnwidth]{figures/ablation_study/examples/input5.png}
    %     \caption{Input image}
    % \end{subfigure}
    % \begin{subfigure}{0.24\columnwidth}
    %     \centering
    %     \includegraphics[width=\columnwidth]{figures/ablation_study/examples/anno5.png}
    %     \caption{Ground truth}
    % \end{subfigure}
    % \begin{subfigure}{0.24\columnwidth}
    %     \centering
    %     \includegraphics[width=\columnwidth]{figures/ablation_study/examples/prediction5.png}
    %     \caption{Segmentation}
    % \end{subfigure}
    % \begin{subfigure}{0.24\columnwidth}
    %     \centering
    %     \includegraphics[width=\columnwidth]{figures/ablation_study/examples/ep_uncertainty5.png}
    %     \caption{Model unc.}
    % \end{subfigure}

\caption{Qualitative results with Bayesian ERFNet (all, $p = 50 \%$) trained on the ISPRS Potsdam dataset~\citep{Potsdam2018}. High model uncertainty in misclassified regions validates that our Bayesian ERFNet provides consistent uncertainty estimates.}
\label{F:results_ablation_examples}
\end{figure}

\cref{T:results_ablation_model} summarises our results. With highest accuracy and \ac{mIoU}, Bayesian ERFNet-All, proposed in \cref{SS:network_architecture}, trained with $p = 50 \%$ performs best.
%This indicates that a fully Bayesian ERFNet provides a high-quality probabilistic model at test time, while benefiting from strong regularisation at train time in aerial semantic mapping tasks with ERFNet. 
Noticeably, our Bayesian ERFNet-All ($p = 50 \%$) outperforms its strongest non-Bayesian counterpart (standard, $p = 50 \%$) by $3.8 \%$ \ac{mIoU} while resulting in $8.3 \%$ improved \ac{ECE}. Qualitatively, \cref{F:results_ablation_examples} confirms high uncertainty in misclassified, cluttered regions, hence providing a reliable acquisition function for \ac{AL} (\cref{eq:mutual_information}). These results validate that our probabilistic model interpretation via \ac{MC} dropout improves performance and provides reliable uncertainty estimates.

\begin{figure}[!h]
    \includegraphics[width=\linewidth]{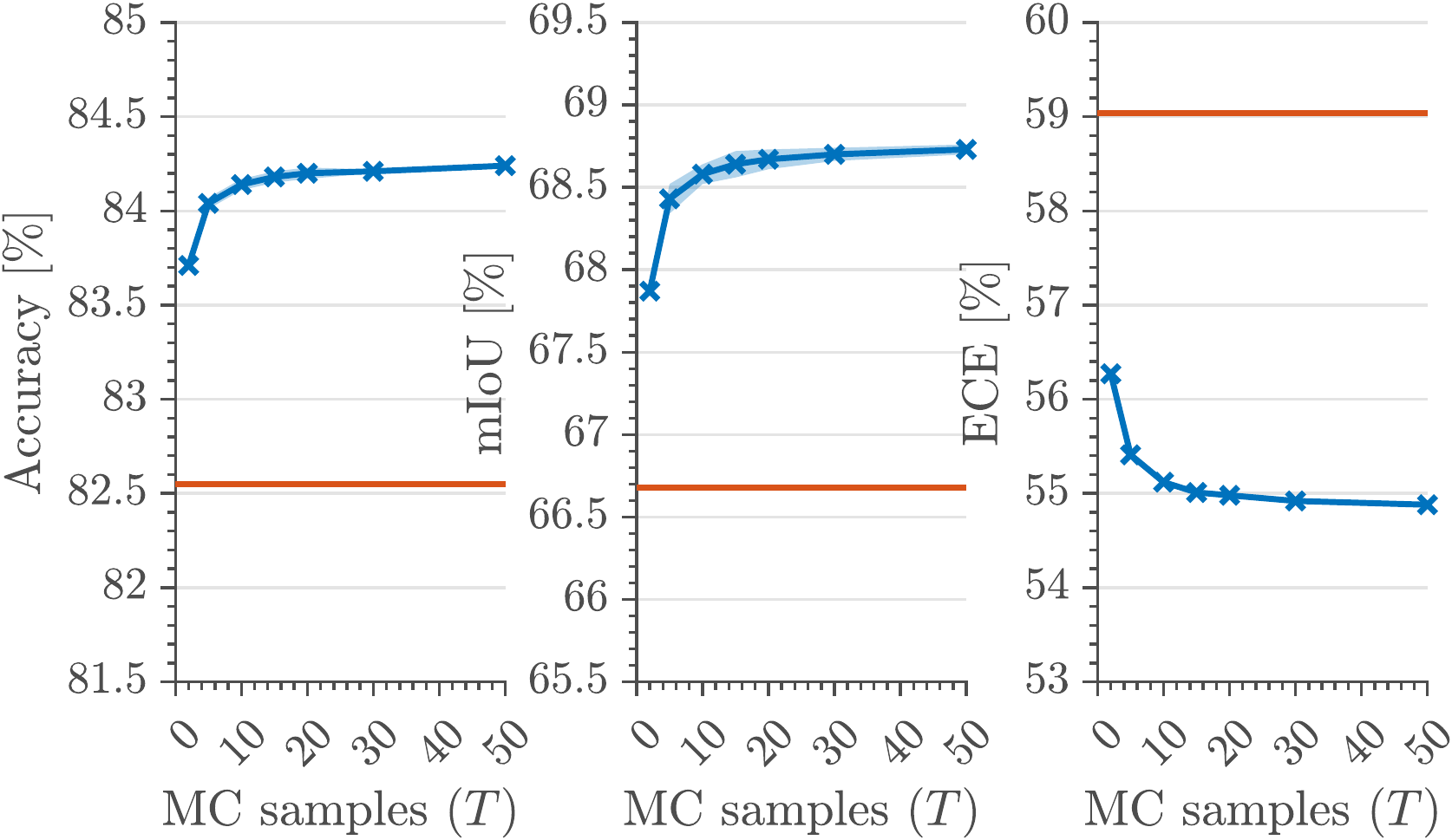}
    \caption{Our Bayesian ERFNet (all, $p = 50\%$, blue) with varying \ac{MC} dropout samples compared against non-Bayesian ERFNet (all, $p = 50\%$) (orange) on the ISPRS Potsdam dataset~\citep{Potsdam2018}. Metrics are averaged over three trials with shaded regions indicating standard deviations. For $T = 50$, Bayesian ERFNet improves mIoU by $3.1 \%$ (left, middle) and reduces ECE by $7.6 \%$ (right).}
    \label{F:results_ablation_dropout_samples}
\end{figure}

\begin{figure*}[!h]
    \centering
    \begin{subfigure}{0.32\textwidth}
        \includegraphics[width=0.49\textwidth]{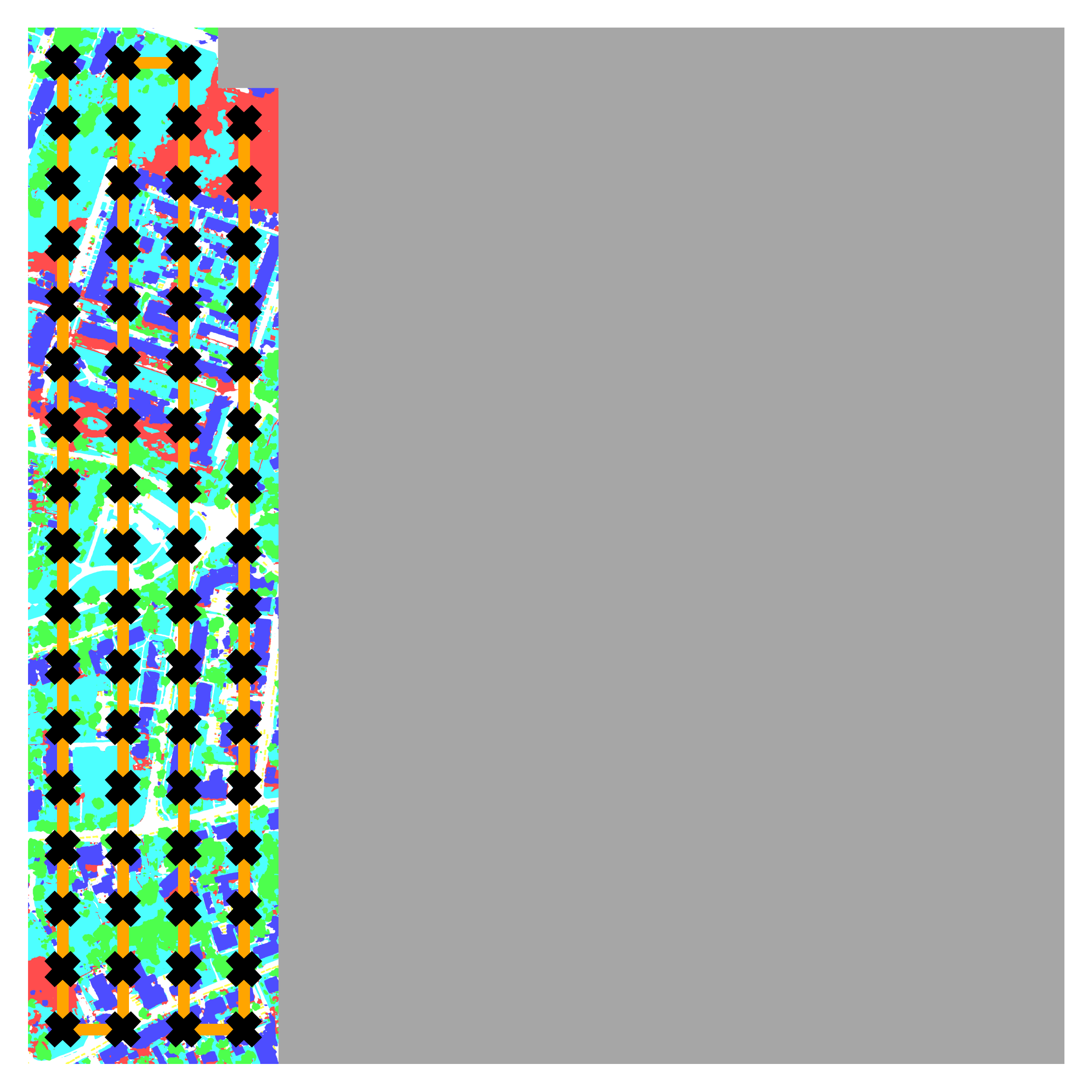} \hfill
        \includegraphics[width=0.49\textwidth]{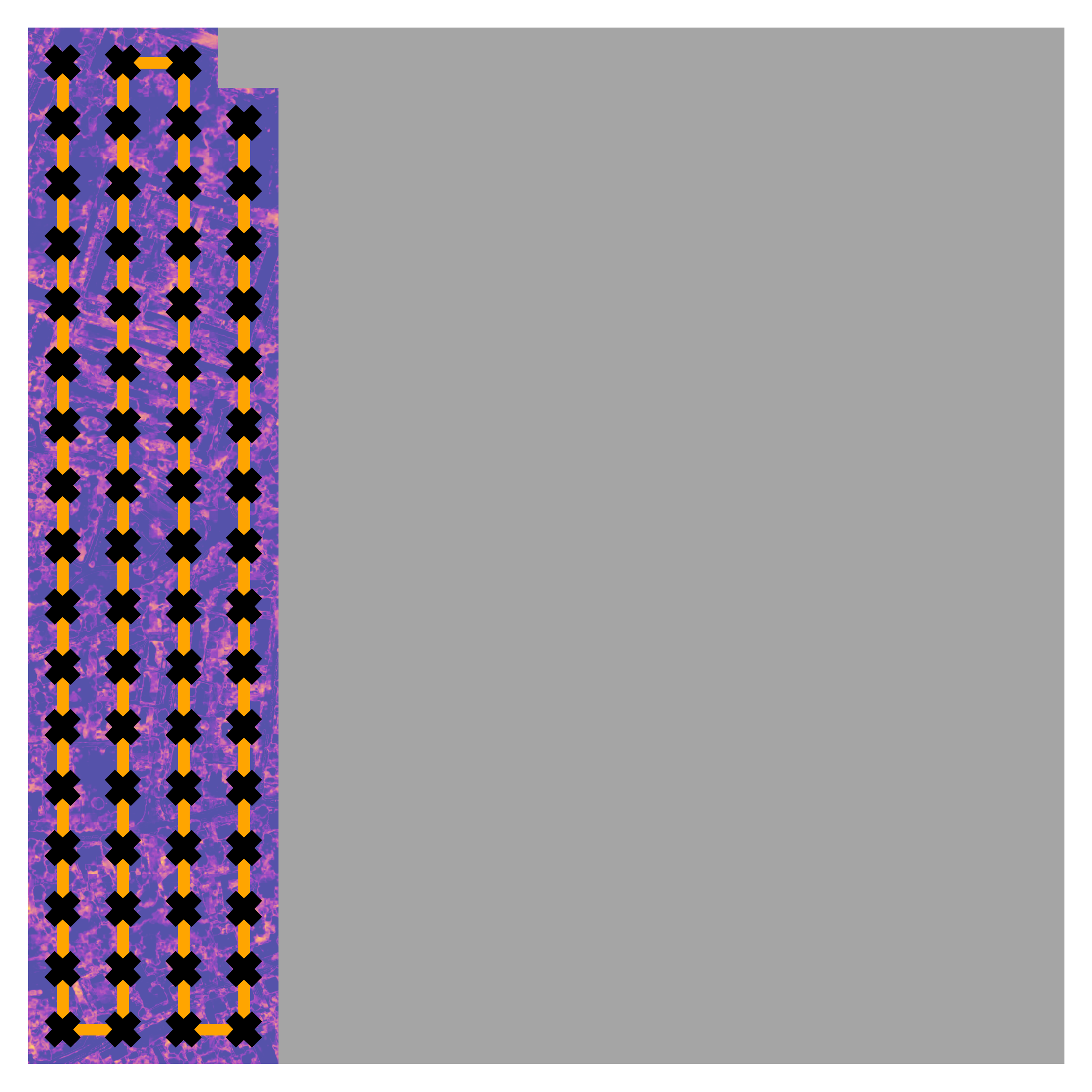}
        \caption{Lawnmower} \label{SF:result_lawnmower}
    \end{subfigure} 
    \begin{subfigure}{0.32\textwidth}
        \includegraphics[width=0.49\textwidth]{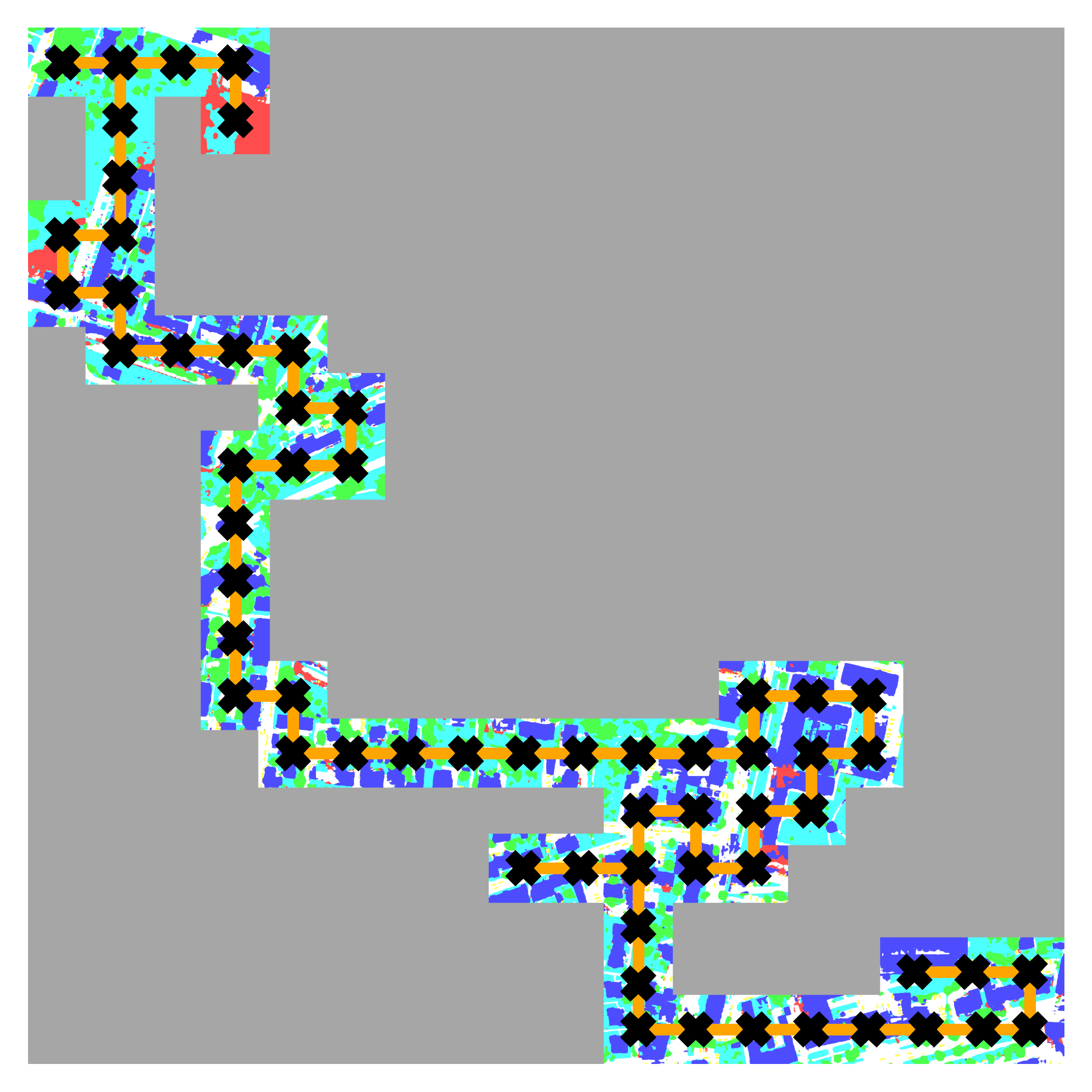} \hfill
        \includegraphics[width=0.49\textwidth]{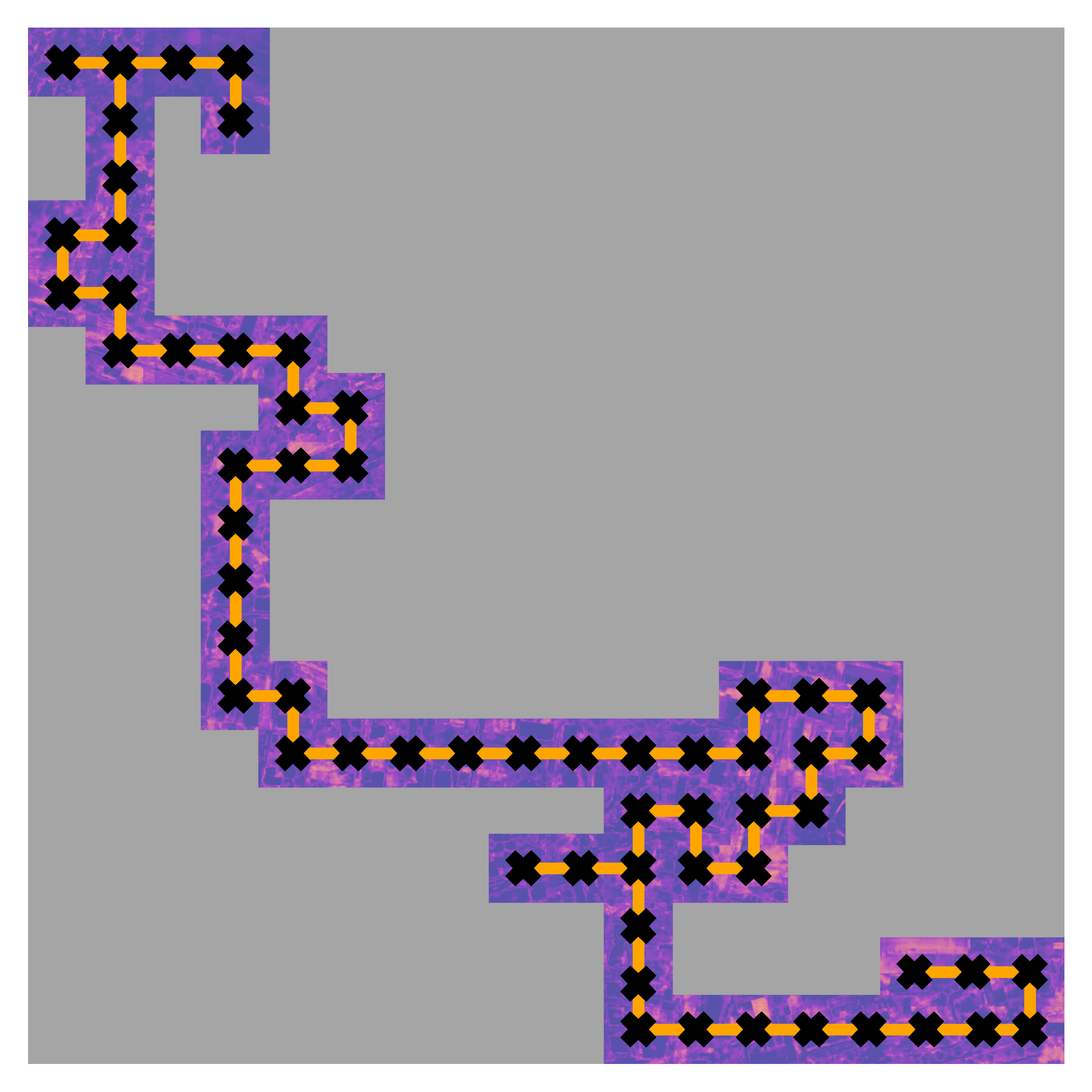}
        \caption{Image-based} \label{SF:result_image}
    \end{subfigure}
    \begin{subfigure}{0.32\textwidth}
        \includegraphics[width=0.49\textwidth]{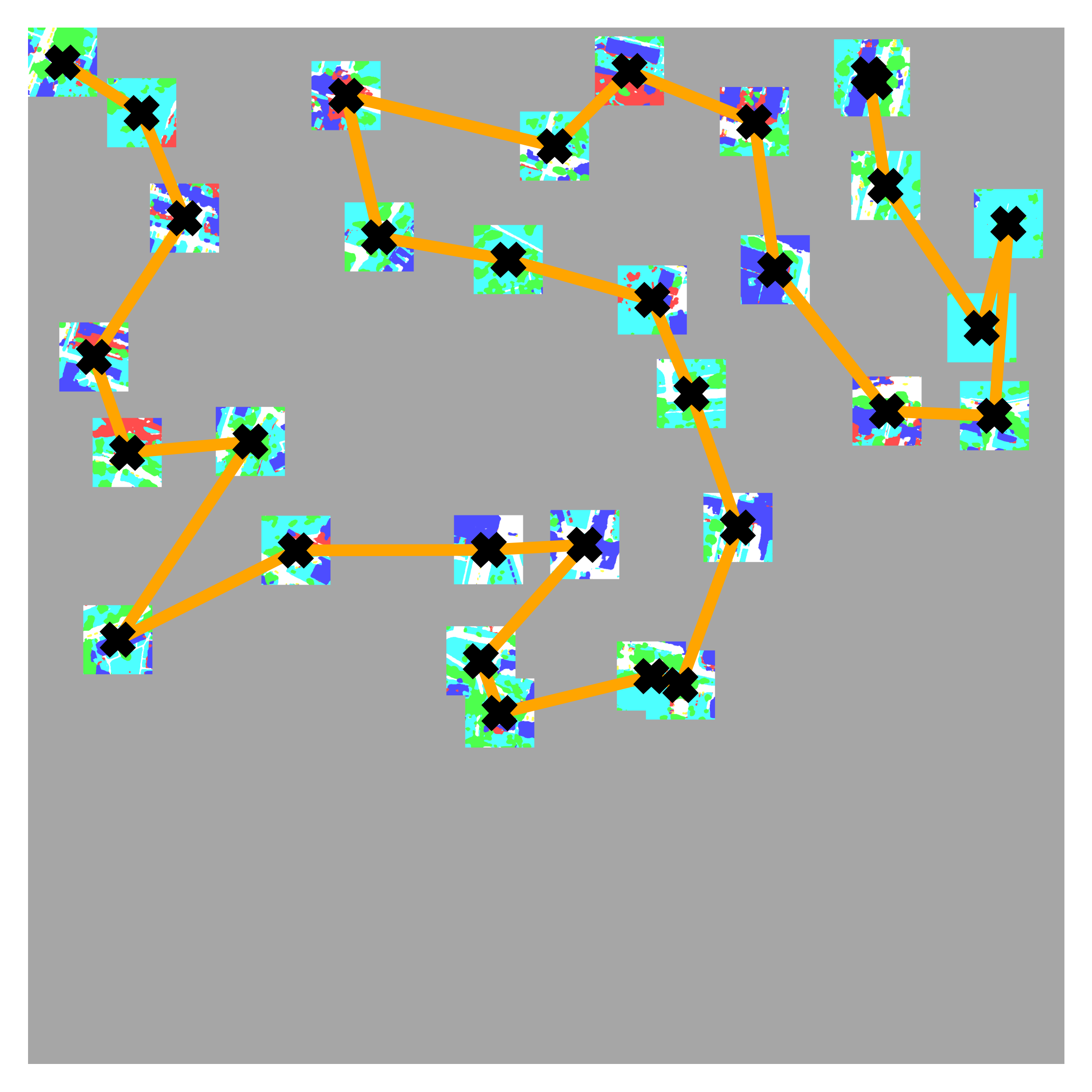} \hfill
        \includegraphics[width=0.49\textwidth]{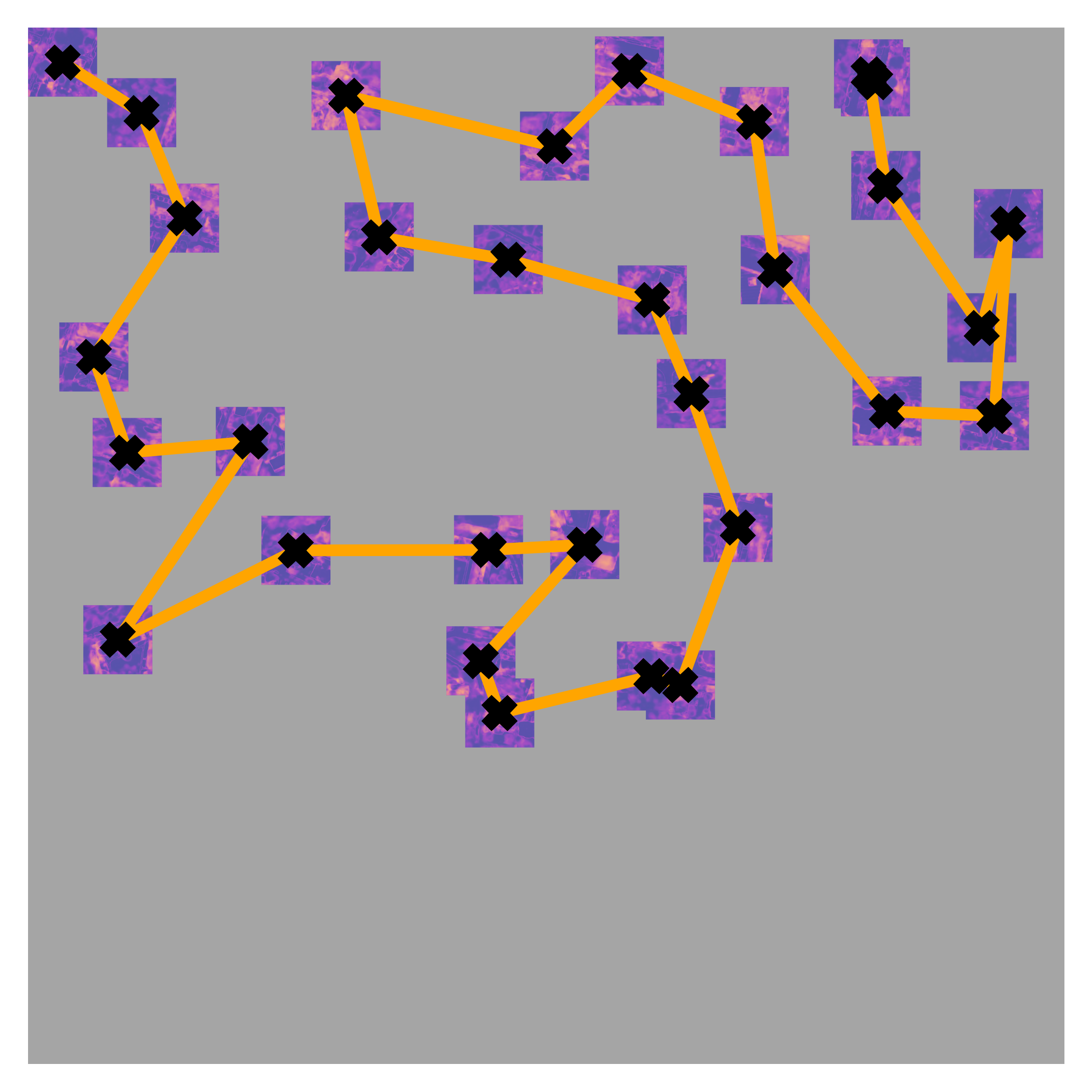}
        \caption{Fixed-horizon} \label{SF:result_horizon}
    \end{subfigure}
    \caption{Qualitative mapping and planning results for one mission with (a) the coverage baseline, (b) image-based, and (c) fixed-horizon strategy on the ISPRS Potsdam orthomosaic~\cite{Potsdam2018}. Orange lines show planned paths with black crosses indicating collected training images. Semantic and model uncertainty maps are shown, where grey areas depict unexplored terrain. Our planners in (b)-(c) adaptively plan to maximise the information value of collected images leading to superior \ac{AL} performance over the coverage pattern in (a).}
    \label{F:potstam_results_maps}
\end{figure*}

% One drawback of \ac{MC} dropout is its need to run $T$ complete forward passes through the network. 
%Although the model uncertainty quantification is DL model-agnostic, we stress the fact that our base architecture, ERFNet, is deliberately chosen for its real-time semantic segmentation capabilities enabling us to run multiple forward passes while maintaining moderate inference speeds at planning time. 
To assess computational requirements and support the applicability of our method for online replanning, we study the performance of our Bayesian ERFNet with varying numbers of \ac{MC} dropout samples $T = \{2,\,5,\,10,\,15,\,20,\,30,\,50\}$ in \cref{F:results_ablation_dropout_samples}. As the number of MC dropout samples increases, segmentation performance and \ac{ECE} both improve. Favourably for active robotic decision-making, $T \approx 20$ samples are already sufficient for converging performance gains.

\subsection{Planning for Active Learning Evaluation} \label{SS:planning_evaluation}

The following results on planning for \ac{AL} suggest that our online planning framework reduces the number of labelled images needed to maximise segmentation performance compared to static coverage paths. Moreover, we show that fixed-horizon planning with a lookahead is better for \ac{AL} performance than a greedy, one-step approach.

These claims are evaluated on the ISPRS Potsdam orthomosaic~\citep{Potsdam2018}. As in \cref{SS:ablation_study}, we simulate images and labels from $30$\,m altitude % with a square footprint, downwards-facing camera, and \ac{GSD} of $15 ~ cm/px$. We 
to create 1000 validation and 3500 test datapoints. Data collection missions are executed in the disjoint $900$\,m$\times900$\,m training area shown in \cref{F:potsdam_dataset}. Our proposed Bayesian ERFNet (all, $p = 50 \%$) is pretrained on the Cityscapes dataset~\citep{Cordts2016}, so that each training starts from the same checkpoint. Each experiment considers $10$ subsequent \ac{UAV} data collection missions with a flight budget of $B = 1800$\,s per mission, starting at $(30,\,30,\,30)$\,m. After each mission, we retrain the model until convergence with batch size $8$ and weight decay $\lambda = (1 - p) / 2N$, where $p = 0.5$ and $N$ is the number of collected images up to the current mission. After each mission, the model is reset to its pretrained checkpoint to avoid catastrophic forgetting and other effects from simply accumulating train time.

\begin{figure}[!t]
    \centering
    \includegraphics[width=0.96\linewidth]{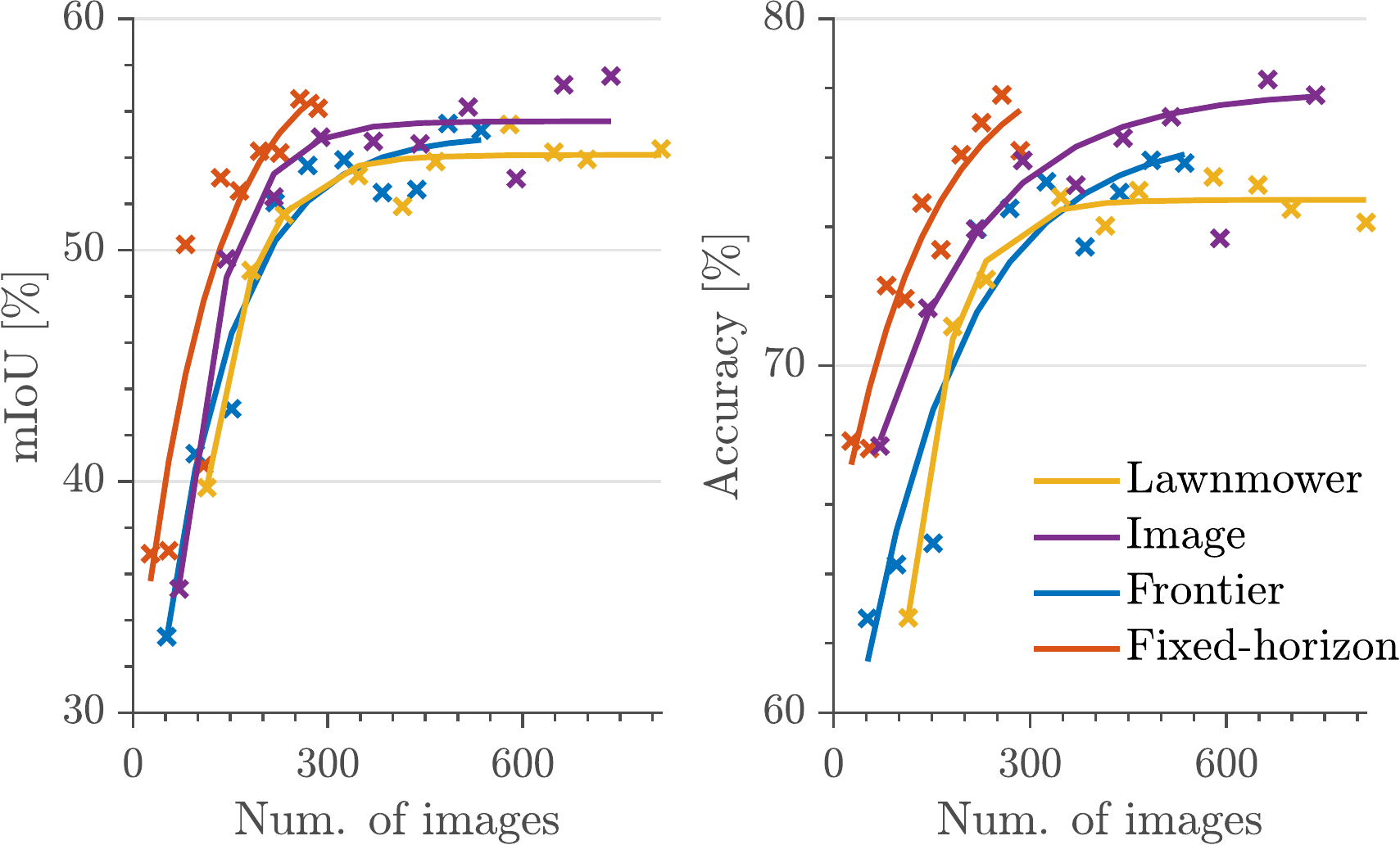}
    \caption{Comparison of \ac{AL} performance of our planners on the ISPRS Potsdam orthomosaic~\citep{Potsdam2018}.
    %\ac{AL} performance is reflected by maximising accuracy and \ac{mIoU} with fewer labelled images. 
    Steeper curves indicate better performance.
    Our informative planners either outperform or perform on par with the strong coverage baseline (yellow). Our fixed-horizon strategy (orange) performs best as it can plan multiple steps ahead based on the uncertainty map.}
    \label{F:potstam_al_results_plot}
\end{figure}

To verify our claims, we compare \ac{AL} performance of the planning methods introduced in \cref{SS:path_planning} to commonly used static coverage paths. We assess segmentation performance by pixel-wise accuracy and \ac{mIoU} over the number of training samples after each retraining. For a fair comparison, we implement a coverage baseline by changing path orientations and distances between measurement positions in the missions to maximise both spatial coverage and training data diversity. In our fixed-horizon planner, we use a lookahead of $N = 5$. Following \citet{Popovic2020}, we apply the Covariance Matrix Adaptation Evolution Strategy~\citep{hansen2001completely} optimisation routine % since it outperforms other optimisation procedures in informative path planning for terrain monitoring 
and tune its hyperparameters to trade-off between runtime and path quality. The frontier-based planner uses a candidate sampling distance of $15$\,m, and the image-based planner uses a step size of $50$\,m and an edge width of $10$\,px.

\cref{F:potstam_al_results_plot} summarises the planning results. Better \ac{AL} performance is reflected by maximising accuracy and \ac{mIoU} with less acquired images that must be labelled by a human operator. As indicated by the steeper-rising curves, our proposed fixed-horizon (orange) and local image-based (purple) strategies clearly outperform the strong coverage baseline (yellow), since they allow active decision-making for data collection. %as they reach higher accuracy and \ac{mIoU} on the test set with equal number of or even less labelled training images.
Frontier-based planning (blue) performs on par with the coverage baseline.
%Qualitatively, \cref{F:potstam_results_maps} confirms that our approach adaptively plans to maximise information value (model uncertainty) of collected images improving \ac{AL} performance over non-adaptive spatial coverage paths.
Qualitatively, \cref{F:potstam_results_maps} confirms the adaptive behaviour of our approaches, which plan paths to target regions of high model uncertainty and gather images about different environment features.
These results verify that our online planning framework effectively reduces human labelling effort while maximising segmentation performance.% especially when leveraging the image-based and fixed-horizon lookahead strategies.

Particularly, \cref{F:potstam_al_results_plot} shows that our fixed-horizon with lookahead strategy performs better for \ac{AL} than the two greedy one-step methods. The fixed-horizon planner already reaches $\sim 57\%$ \ac{mIoU} and $\sim 78\%$ accuracy with $\sim 250$ labelled images while the second-best local image-based strategy requires $\sim 520$ images, i.e. around twice the number of training samples, to achieve these levels.
%The greedy frontier-based and the static coverage planners show even weaker labelling efficiency.
The superior performance of the fixed-horizon method demonstrates the benefits of planning multiple steps ahead in our integrated framework to achieve efficient training data acquisition.

\section{Conclusions and Future Work} \label{S:conclusions}
This paper introduced a general informative path planning framework for active learning in aerial semantic mapping scenarios.
Our framework exploits a Bayesian \ac{FCN} to estimate model uncertainty in semantic segmentation.
During a mission, the semantic predictions and model uncertainty are used as an input for terrain mapping.
A key aspect of our work is that we link the mapped model uncertainty to a planning objective guiding \ac{UAV} to collect the most informative images for training the \ac{FCN}.
This pipeline reduces the total number of images that must be labelled to maximise segmentation performance, thus conserving human effort, time, and cost.

We validated our Bayesian \ac{FCN} in an ablation study, showing that it both improves segmentation performance compared to a non-Bayesian baseline and provides consistent model uncertainty estimates as a basis for \ac{AL}. The integrated system for informative planning with the \ac{FCN} was evaluated in an urban mapping scenario using real-world data. Results show that our framework effectively reduces labelled training data requirements compared to static coverage paths. Moreover, we demonstrate that planning with a lookahead strategy in our approach improves data-gathering efficiency compared to simpler greedy one-step planners.
Future work will investigate path planning at different altitudes and exploiting semantic map information to target specific informative classes, e.g. buildings or cars.

\section*{Acknowledgement}
We would like to thank Matteo Sodano and Tiziano Guadagnino for help with our experiments and proofreading. We would like to thank Jan Weyler for providing a PyTorch Lightning implementation of ERFNet.

\bibliographystyle{IEEEtranN}
\footnotesize
\bibliography{2022-iros-rueckin}

\end{document}